%% file: main.tex
\documentclass[11pt]{article}
\usepackage[final]{acl}
\makeatletter
\ifdefined\IfFormatAtLeastTF
  \IfFormatAtLeastTF{2025-06-01}{%
    \@ifpackagelater{lineno}{2026/02/17}{}{%
      \@ifpackagelater{lineno}{2026/02/15}{%
        \RemoveFromHook{build/column/before}[lineno]%
        \AddToHook{build/column/before}[lineno]{%
          \protected@write\@auxout{}{%
            \string\def\string\@LN@column{%
              \if@firstcolumn2\else1\fi}}}%
      }{}%
    }%
  }{}%
\fi
\makeatother
\usepackage{times}
\usepackage{latexsym}
\usepackage[T1]{fontenc}
\usepackage[utf8]{inputenc}
\usepackage{microtype}
\usepackage{inconsolata}
\usepackage{graphicx}
\usepackage{booktabs}\usepackage{longtable}
\usepackage{amsmath,amssymb}
\usepackage{multirow}
\usepackage{xcolor}
\newcommand{\heldout}{held-out}

\usepackage{enumitem}
\usepackage{fontawesome5}
\title{From Copying to Composition: \\How Language Models Develop In-Context Learning}

\author{
Minglai Yang$^{1}$\thanks{Corresponding author.}\;
Xinyu Guo$^{2}$\;
Renxiong Wang$^{1}$\;
Sunjie Hou$^{1}$\;
Daniel Yue Zhang$^{1}$\;
Yunzhong He$^{1}$\\[0.3em]
$^{1}$Scale AI\quad
$^{2}$University of Arizona\\[0.3em]
\texttt{minglai.yang@scale.com}
}

\begin{document}
\maketitle
\input{sections/00_abstract}
\begin{figure*}[t]
\centering
\includegraphics[width=0.95\textwidth]{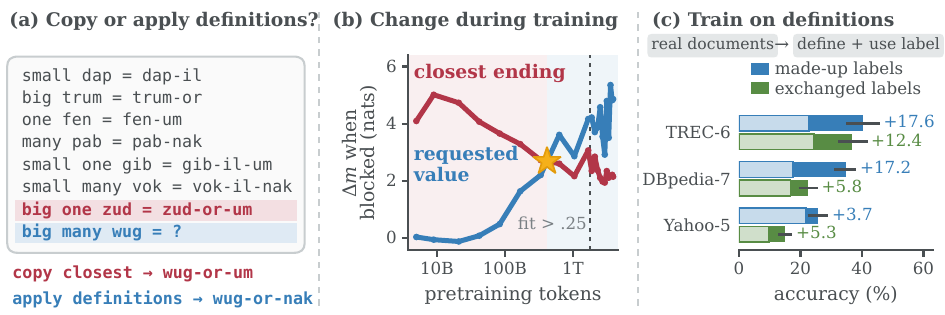}
\caption{Study overview. (a) The task makes copying and applying prompt definitions predict different answers. (b) Across pretraining checkpoints (OLMo-2-7B shown), the requested value overtakes the closest ending as the source of the answer (star) long before accuracy crosses chance (dashed line). (c) Fine-tuning SmolLM2-1.7B on definition exercises rewritten from real documents raises accuracy on three unseen datasets with made-up or exchanged labels (pale bars: a matched fine-tune on the original documents, solid extensions: the gain from the exercises).}
\label{fig:overview}
\end{figure*}
\input{sections/01_intro}
\input{sections/06_related}
\input{sections/02_task}
\input{sections/05_reading}
\input{sections/04_dynamics}
\input{sections/06_data}
\input{sections/07_discussion}
\input{sections/07_conclusion}
\clearpage
\input{sections/08_limitations}
\input{sections/08b_ethics}

\bibliography{references}
\clearpage
\appendix
\input{sections/09_appendix}
\end{document}

%% file: sections/00_abstract.tex
\begin{abstract}
In-context learning (ICL) allows language models to adapt their behavior from instructions and demonstrations in a prompt, but accuracy alone does not reveal how contextual information controls their predictions. We study the acquisition and trainability of ICL with a controlled diagnostic that separates copying a nearby demonstration from integrating information across the prompt. Across 17 models, about a third prefer the copied answer, a few are undecided, and the rest prefer the rule-consistent answer, which goes with higher accuracy and with control by the requested value rather than the closest conflicting example. Checkpoint analyses show that models begin prioritizing task-relevant prompt tokens over conflicting examples long before their accuracy improves beyond chance level. In matched training experiments, synthetic documents that define endings and then query them shift models toward compositional responses on the controlled diagnostic, whereas fine-tuning on definition exercises built from rewritten real documents improves accuracy when prompts reassign label meanings. The results suggest that ICL is not acquired as a single, uniform capability. Instead, prompt control, answer preference, construction without candidates, and transfer can develop along distinct trajectories. 
\end{abstract}

%% file: sections/01_intro.tex
\section{Introduction}

Language models are increasingly expected to adapt at inference time, from instructions, demonstrations, and temporary conventions in the prompt rather than from parameter updates. This ability is commonly described as in-context learning (ICL). We know much less about how this capacity is acquired during training than about how well trained models display it. Understanding ICL acquisition requires tracking how models shift their reliance on contextual information throughout training. Knowing when accuracy improves is not enough.

The same answer can result from different uses of the context: a model may reuse the output pattern of a demonstration similar to the query, or integrate mappings distributed across the prompt and apply them to a new combination. In ordinary ICL evaluations these behaviors predict the same output, so correctness shows that the context was useful but not how it was used. Studying the acquisition of ICL therefore needs a diagnostic in which different ways of using the prompt make different predictions.

These observations motivate a developmental question that final-answer accuracy alone cannot resolve: how does a model's use of different sources of contextual information change during training? We track four aspects of ICL that, as we show, develop on different schedules: which prompt source controls the answer, which answer the model prefers, whether it can construct that answer without candidates, and whether a trained change transfers to other evaluations.

To make these aspects distinguishable, we construct a controlled task inspired by work on compositional generalization and made-up-word tests \citep{hupkes2020compositionality,berko1958wug}. Each prompt defines endings for two binary attributes, demonstrates three of the four possible combinations on made-up words, and asks for the missing combination on a new word (Figure~\ref{fig:overview}). Combining the two prompt-defined mappings yields the rule-consistent (correct) answer, whereas reusing the conflicting ending from the demonstrated combination that shares the first requested attribute yields a distinct copied answer. Because the assignments change in every prompt, the task cannot be solved using a fixed word--ending mapping. We measure the copied answer's probability normalized over these two candidates, denoting it by $P(\mathrm{copy})$. Interventions on the conflicting ending in the closest example or on the requested value in the question measure which source controls the candidate-score difference, and activation patching \citep{heimersheim2024patching} and the disabling of selected attention heads \citep{wang2023ioi} trace that control to internal states and heads.

Across 17 models, the diagnostic sorts answer preferences from copying the nearest demonstration to producing the correct form, and the composing models score higher and generate the form without candidates. Across pretraining checkpoints, sensitivity to the requested value overtakes sensitivity to the conflicting example hundreds of billions of tokens before accuracy rises above chance. Most other checkpoint series that reach composition show the same ordering, OLMo-3 and Comma on every criterion (Section~\ref{sec:dynamics}).

Matched continued-training experiments then show which training signals change these behaviors and where the improvements transfer. Training on synthetic documents that define endings and then query them reverses the preferred answer on the controlled task and lets models generate the missing form. Fine-tuning on definition exercises built from rewritten real documents improves performance on evaluations that assign new meanings to labels. Each signal transfers the use of the prompt it rewards, so the exercises leave the controlled task unchanged.

Our work makes three contributions. First, we introduce a controlled diagnostic in which two ways of using the context predict different answers. Second, we measure answer preference and prompt dependence across 17 models and multiple checkpoint series showing that prompt control shifts long before accuracy crosses chance. Third, matched continued-training experiments show that the compositional answer must be built rather than unblocked: suppressing copy-favoring heads shifts the balance between candidates, training on synthetic documents that define endings and then query them creates the missing answer, and each signal transfers the use of the prompt it rehearses (Section~\ref{sec:lever}).

%% file: sections/06_related.tex
\section{Related work}
\label{sec:related}

\paragraph{In-context learning and composition.}
Accounts of in-context learning include copying through induction-like mechanisms \citep{olsson2022induction,bietti2023birth,reddy2024mechanistic,akyurek2024icll}, inference over a latent task, and an update during the forward pass \citep{xie2022bayesian,vonoswald2023icl}. ICL behavior varies with training distribution and stage \citep{chan2022data,singh2023transient}. Controlled tasks show strategies between repetition and exact rule use \citep{niu2025illusion,bhattamishra2024understanding}, and models can recognize a task without learning it \citep{yang2025trtl,kim2025schema}. Our task connects this issue to compositional generalization and made-up-word tests \citep{hupkes2020compositionality,lake2018generalization,kim2020cogs,lake2023human,berko1958wug,weissweiler2023wugs}, and to work on prompt-defined labels \citep{wei2023symbol,wei2023larger}. Morphology offers a long-studied case of the same contrast between applying a rule and analogizing to stored forms \citep{pinker1988language,albright2003rules}, and recent work asks which of the two language models do \citep{hofmann2025analogy,anh2024wug,ismayilzada2024morphcomp}. Tokenization also shapes what counts as a morpheme for a model \citep{sennrich2016neural,rust2021tokenizer}, which our Quechua results bear out. We make copying a similar example and applying the supplied mappings predict different answers, then compare their scores and intervene on the supporting prompt tokens. Attention weights provide incomplete explanations of model behavior \citep{jain2019attention,serrano2019attention}. Our interventions directly test which prompt tokens change the candidate scores.

\paragraph{Mechanisms of context use.}
Mechanistic studies locate parts of in-context behavior in attention heads. Induction and retrieval heads copy from the context \citep{olsson2022induction,wu2024retrieval}, a copy-suppression head counteracts them \citep{mcdougall2023copy}, token-level and concept-level copying travel separate routes \citep{feucht2025dualroute}, and a set of demonstrations can be summarized into a task or function vector \citep{hendel2023taskvectors,todd2024function}. Our interventions follow this tradition of patching and ablation \citep{geiger2021causal,meng2022locating,goldowskydill2023path,wu2023interpretability}, but we ask a different question: which prompt span controls the answer at each stage of training.

\paragraph{Development during training.}
Checkpoint studies usually measure abilities by accuracy gains \citep{chen2024sudden,evanson2023stages,tigges2024consistent,arora2024causalgym}. Work on combined abilities finds that they emerge after their components \citep{liu2026curriculum}, while other studies track reasoning, arithmetic, grammar, and internal components across training \citep{ye2025implicit,zhao2026shattered,chang2022word,pan2026blackbox}. We instead track copied-versus-correct scores and prompt-token dependence at every checkpoint. We extend work that localizes or modifies mechanisms for context use \citep{yin2024lofit,prakash2024finetuning} to the training trajectory.

%% file: sections/02_task.tex
\section{Task and measurements}
\label{sec:task}

\paragraph{Task construction.} Figure~\ref{fig:overview}a shows one prompt. Each prompt assigns a fresh ending to each of four feature values, shows three of the four value combinations, and asks for the missing combination on a new stem. In the example, applying the definitions gives \textit{wug-or-nak}, whereas copying the closest example, \textit{big one zud = zud-or-um}, gives \textit{wug-or-um}. The words, endings, and mappings are sampled anew for every prompt. A second scheme uses short codes without a made-up word, and a reverse-order version serves as a control (Appendices~\ref{app:prereg} and~\ref{app:controls}).

\paragraph{Formalization and scores.}
Let $\mathcal U=\{u_0,u_1\}$ and $\mathcal V=\{v_0,v_1\}$ be the feature domains. Each prompt defines a map $\phi:\mathcal U\sqcup\mathcal V\rightarrow\mathcal E$ from their four values to four endings, with
\begin{equation}
 f_\phi(s,u,v)=s\mathbin\Vert\phi(u)\mathbin\Vert\phi(v),
\label{eq:composition}
\end{equation}
where $\Vert$ denotes concatenation. Equation~\ref{eq:composition} composes the two prompt-defined endings in feature order. The query requests a held-out pair $(u^\star,v^\star)$ on a new stem $s^\star$. If $\bar v^\star$ is the other value in $\mathcal V$, the closest example supplies $(u^\star,\bar v^\star)$. In Figure~\ref{fig:overview}a, $u^\star=\textit{big}$, $v^\star=\textit{many}$, and $\bar v^\star=\textit{one}$. The rule-consistent and copied candidates are therefore
\begin{equation}
 y_r=f_\phi(s^\star,u^\star,v^\star),\qquad
 y_c=f_\phi(s^\star,u^\star,\bar v^\star).
\label{eq:candidates}
\end{equation}
The candidates in Equation~\ref{eq:candidates} share the stem and first ending and differ only where copying conflicts with the requested value.

For prompt $x$, let $p_r=p_\theta(y_r\mid x)$ and $p_c=p_\theta(y_c\mid x)$. We measure their competition by
\begin{equation}
\begin{aligned}
 m&=\log p_r-\log p_c,\\
 P(\mathrm{copy})&=\frac{p_c}{p_c+p_r}=\sigma(-m).
\end{aligned}
\label{eq:competition}
\end{equation}
Positive $m$, equivalently $P(\mathrm{copy})<.5$, favors the correct form. Four-choice accuracy additionally requires $y_r$ to outrank all four combinations, while exact generation supplies no candidates.

\paragraph{Blocking and the switch of control.}
We block attention from the answer position to selected tokens without deleting them \citep{geva2023dissecting} (Figure~\ref{fig:attention}a). Let $K_S$ denote this intervention for span $S$ and $\Delta_Sm=m_\theta(K_Sx)-m_\theta(x)$. For the closest ending $S_e$, the conflicting ending in the closest example (\textit{-um}), and the requested value $S_q$ (\textit{many}), define
\begin{equation}
\begin{aligned}
 d_{i,\tau}&=|\Delta_{S_e}m_{i,\tau}|-|\Delta_{S_q}m_{i,\tau}|,\\
 D_\tau&=\frac{1}{n}\sum_{i=1}^n d_{i,\tau}.
\end{aligned}
\label{eq:handoff}
\end{equation}
Positive $D_\tau$ in Equation~\ref{eq:handoff} indicates greater control by the closest ending, and negative $D_\tau$ greater control by the requested value. Across checkpoints $\tau_1<\cdots<\tau_J$, the switch of control is the first adjacent bracket $[\tau_{j-1},\tau_j]$ in which mean $D$ changes from positive to negative. Other prompt lines and unrelated tokens serve as blocking controls.

\paragraph{Activation patching at the word naming the requested value.} We run a second prompt that asks for the other value of the same attribute. Following activation-patching methods \citep{heimersheim2024patching}, after layer~4 we replace the original run's internal representation at that word with the representation from the second run, then let the remaining layers continue. We score success strictly: the model must write the full form required by the replacement value. Targets shown in the prompt are reported separately from never-shown ones, since never-shown targets alone require constructing a new form (Figure~\ref{fig:patch}; layer choice: Appendices~\ref{app:das} and~\ref{app:transfer}).

\paragraph{Models.} We test released Pythia and OLMo models \citep{biderman2023pythia,olmo2025olmo2} and Qwen and Gemma families \citep{qwen3,qwen25,gemma3}. To follow changes during training, we use released intermediate checkpoints and public training trajectories \citep{groeneveld2024olmo,muennighoff2024olmoe,liu2023llm360}. Continued-training experiments restart intermediate checkpoints of SmolLM2-360M and SmolLM2-1.7B \citep{allal2025smollm2} on a public proxy of their corpus (Appendices~\ref{app:part7b} and~\ref{app:part10}). Definition-exercise fine-tunes use SmolLM2-1.7B and SmolLM3-3B \citep{huggingface2025smollm3} (Appendix~\ref{app:transfer}). Low-rank adapters are trained on five released 7--8B bases (Qwen2.5-7B, OLMo-2-7B, Llama-3.1-8B, Comma-v0.1-1T, RedPajama-INCITE-7B; Figure~\ref{fig:lora_boundary}). A direct task fine-tune uses Pythia-1.4B (Appendix~\ref{app:induce}). Code-trained Qwen variants \citep{hui2024qwen25coder} serve one control (Appendix~\ref{app:controls}).

%% file: sections/05_reading.tex
\section{Answer preference reflects prompt control}
\label{sec:fingerprint}
\label{sec:routes}
\label{sec:heads}
\label{sec:reading}

\begin{figure}[t]
\centering
\includegraphics[width=0.95\columnwidth]{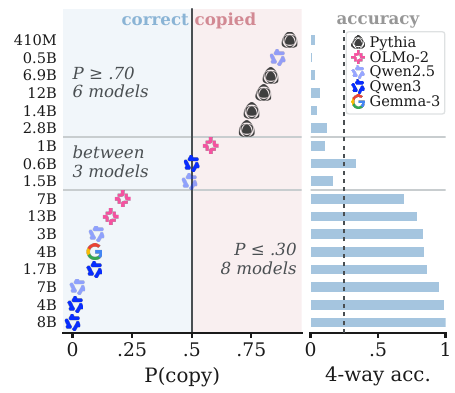}
\caption{Copied-answer probability (left) and four-choice accuracy (right) for 17 models, grouped by the $.70$ and $.30$ thresholds; marker shape gives the family (full scores in Table~\ref{tab:fingerprint}).}
\label{fig:fingerprint}
\end{figure}

\begin{figure}[t]
\centering
\includegraphics[width=0.95\columnwidth]{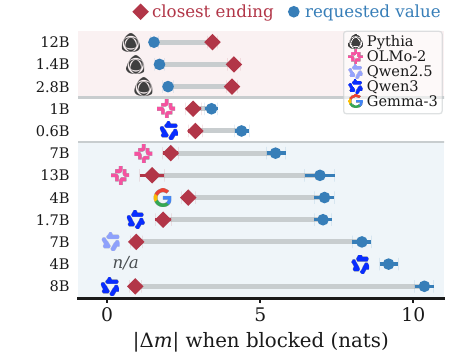}
\caption{Change in $m$ after blocking the closest ending or the requested value, model by model (all blocked spans: Appendix~\ref{app:tables}).}
\label{fig:knockout}
\end{figure}

\begin{figure}[t]
\centering
\includegraphics[width=0.8\columnwidth]{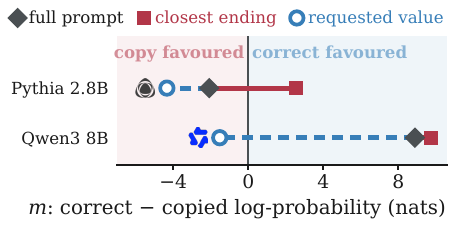}
\caption{Blocking a single prompt span reverses the preferred answer: mean $m$ with the full prompt and after blocking each span, for high-copy Pythia-2.8B and low-copy Qwen3-8B.}
\label{fig:mech}
\end{figure}

\begin{figure*}[t]
\centering
\includegraphics[width=0.93\textwidth]{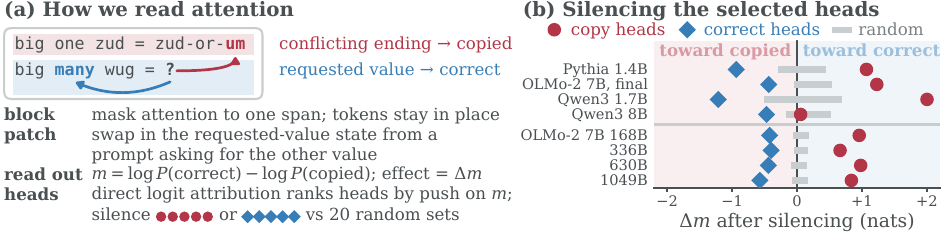}
\caption{Attention analysis. (a) The answer position reads two prompt spans, the conflicting ending of the closest example (favors the copied answer) and the requested value (favors the correct one). We block attention to a span, patch the requested-value state from a prompt asking for the other value, and read out the change in $m$. Direct logit attribution picks the five heads pushing hardest toward each answer, which are silenced and compared with random sets. (b) Blocking a span moves $m$ far more than silencing either set (Figure~\ref{fig:knockout}).}
\label{fig:attention}
\end{figure*}

\textbf{Three groups of models.} The models spread across the copying-to-composition range. Six assign at least $.70$ of the copied and correct candidates' combined probability to the copied answer, eight assign at most $.30$, and three fall between these thresholds (Figure~\ref{fig:fingerprint}; Table~\ref{tab:fingerprint}). Candidate preference tracks the ability to construct the missing form: the correct answer rarely ranks first in the high-copy group and usually does in the low-copy group, with exact generation following the same ordering. The grouping is not a size effect: Pythia-12B copies while Qwen3-1.7B and Gemma-3-4B compose (Figure~\ref{fig:fingerprint}).

Arbitrary codes and a token-aligned Southern Quechua version preserve the model ordering. Removing the closest combined example selectively weakens the copied candidate, while deleting unrelated lines has little effect (Appendices~\ref{app:instantiations} and~\ref{app:deletion}; Table~\ref{tab:deletion_final}). These controls tie the preference to the competing information in the prompt rather than to wording, suffix inventory, or prompt length (Figure~\ref{fig:controls}).

\paragraph{The controlling prompt source differs between the groups.}
Blocking the closest ending has the larger effect in high-copy models, whereas blocking the requested value has the larger effect in low-copy models (Figure~\ref{fig:knockout}). In the high-copy Pythia models the closest-ending effect is about twice the requested-value effect. In low-copy models the requested-value effect is several times the closest-ending effect. Figure~\ref{fig:mech} shows one model per group (all blocked spans: Appendix~\ref{app:tables}, Table~\ref{tab:knockout}).

The three models between the probability thresholds behave as an intermediate group: their two candidates receive similar probability and, for the models measured, the two effects are much closer than in either of the two main groups (Table~\ref{tab:knockout}), so the balance between the candidates and the controlling span both vary continuously.

\paragraph{Models can read the query without constructing the missing answer.}
Patching the requested-value state redirects every model toward the corresponding combination when that combination already appears in the prompt, but when it requires the unshown combination the Pythia models rarely follow, whereas the other models follow on a substantial share of prompts (Figure~\ref{fig:patch}). High-copy models thus represent enough of the query to produce a shown combination but fail to construct the missing one.

Head-level analyses (Figure~\ref{fig:attention}) place the behavior above the level of single heads. Copy-favoring head sets transfer across prompt samples in several models and lower $P(\mathrm{copy})$ when disabled, yet the preference survives their removal, and no five-head set accounts for most of the token-blocking effect. Blocking, not raw attention weights, tracks the developmental change reported next (Appendix~\ref{app:part5}).

%% file: sections/04_dynamics.tex
\section{Prompt control changes before composition}
\label{sec:development}
\label{sec:dynamics}

\begin{figure}[t]
\centering
\includegraphics[width=0.9\columnwidth]{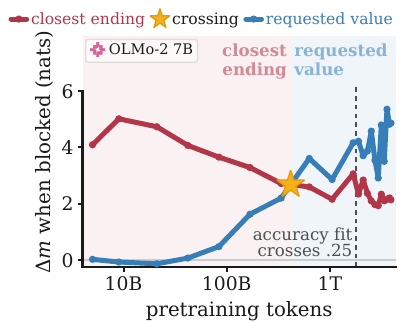}
\caption{Prompt control across pretraining checkpoints, shown for OLMo-2-7B, the most densely sampled series (other series: Appendix~\ref{app:traj}). The requested-value effect overtakes the closest-ending effect between 336B and 630B tokens (star). Four-choice accuracy crosses chance much later (dashed line; Table~\ref{tab:onsets}).}
\label{fig:trajcol}
\end{figure}

The checkpoints confirm what the cross-model contrast predicts: control of the answer changes before the model can construct the missing form.

\textbf{Control switches first.} Early checkpoints are governed by the closest ending. Blocking that ending changes the copied-versus-correct margin far more than blocking the requested value. The balance reverses between 336B and 630B tokens (Figure~\ref{fig:trajcol}; also Figure~\ref{fig:overview}b). At the end of this interval the two candidates receive similar probability and four-choice accuracy has yet to reach chance: prompt control has shifted before the correct answer becomes reliable.

\textbf{Accuracy follows.} Accuracy improves later. The fitted four-choice trajectory crosses chance at about 1.5T tokens on the original items and 1.8T on the re-test items (Table~\ref{tab:onsets}; Appendix~\ref{app:confirm}), well after the two blocking effects cross. The parts of the task are already in place much earlier: the model can copy the requested stem, recover each feature mapping, and generate combinations shown in the prompt hundreds of billions of tokens before it ranks the missing combination first. The long middle stage therefore has every part of the task in place, with composition of those parts the last piece to arrive.

Exact generation of the missing form arrives last, at about 3.6T tokens (Table~\ref{tab:onsets}): parts, then control, then selection, then generation.

\textbf{Other training series.} The ordering recurs in runs from other labs and recipes. Of sixteen released series, Marin-8B, Comma-v0.1-1T, SmolLM3-3B, and OLMo-3-7B end above the accuracy criterion, Apertus is partial, and eleven stay below it. OLMo-3 and Comma satisfy every ordering criterion, shifting hundreds of billions of tokens before their first above-chance checkpoints. SmolLM3 also shifts before accuracy rises, with the closest-example effect later regaining strength. Within the OLMo family, OLMo-2-13B shifts early but briefly reverses, and the older OLMo-7B-0424 run changes sign repeatedly and turns negative for good only after its fitted accuracy onset. Runs that fail to compose never sustain control by the requested value and keep the copied answer. In every run that reaches composition, $D$ first turns negative by the time accuracy crosses chance, and no below-chance run sustains the switch (Appendix~\ref{app:part7a}; Table~\ref{tab:crossings}).

The endpoint comparison links this transition to eventual success: trajectories that exceed chance finish with much larger requested-value effects than those that remain below chance, and in the successful runs $D$ falls below zero before or by the time four-choice accuracy exceeds chance, whereas runs in which the requested value never gains control do not learn the missing combination within the observed training window (Table~\ref{tab:crossings}).

\paragraph{The switch shows at the head level and extends beyond it.} The correct-favoring heads overtake the copy-favoring heads in the same interval (Figure~\ref{fig:heads_traj}). The change is distributed: five heads carry part of the blocking effect, and silencing them moves the balance without creating composition.

The order is a property of pretraining, not of the task: under direct task fine-tuning the blocking crossing and the accuracy rise happen together, or accuracy moves first on a denser grid (Appendix~\ref{app:induce}). The next section asks which training signals produce the switch and what transfers once they do.

%% file: sections/06_data.tex
\section{Training teaches what it practices}
\label{sec:lever}
\label{sec:synthetic}

The checkpoint results show control of the answer passing from the closest example to the requested value. We next ask which training signal produces this change. Synthetic documents that define endings and then query them reverse which prompt source controls the answer and let models produce the missing form. Exposure to definitions and a penalty on the copy-favoring heads do neither, and definition exercises built from rewritten real documents instead teach models to follow the labels a prompt assigns.

\subsection{Practice, not exposure or suppression, moves control}

\textbf{Exposure to definitions.} Definitions embedded in natural text are not the active ingredient. After 1B additional tokens on SmolLM2-1.7B (Appendix~\ref{app:part10}), continued pretraining on mined passages that define and later use a term, on absent-term variants, or on dictionary-prefixed text leaves $D$ positive and exact generation at the plain-continuation level. All four conditions remain controlled by the closest ending (Figure~\ref{fig:p10all_final}).

\textbf{Penalizing copying.} Penalizing the copy-favoring heads changes the balance between the two candidates without creating the missing answer. At SmolLM2-360M, adding to plain continuation a penalty on attention from the answer position to the closest ending, applied to ten heads selected by direct logit attribution for favoring the copied answer, drives those heads' attention to that ending to near zero. With the copy route closed, $D$ stays positive and exact generation stays at zero, while the same penalty on ten random heads leaves both at plain-continuation levels (Figure~\ref{fig:p7b_final}; Appendices~\ref{app:part6} and~\ref{app:part7b}).

\subsection{Synthetic documents reverse copying and transfer what they practice}

\textbf{Synthetic documents.} The synthetic documents reverse both the preferred answer and which prompt source controls it. These documents have the diagnostic's own dictionary-and-query form: each defines endings for a small feature system, shows some combinations, leaves others unshown, and writes out their answers, with nonce stems, markers, and templates drawn from pools disjoint from the evaluation inventories and no code systems (Appendix~\ref{app:part7b}). The suffix diagnostic shares the training format, so the code and Quechua versions serve as transfer probes. Replacing $3\,\%$ of a continued-pretraining stream with these documents moves SmolLM2-1.7B from copy-dominant to query-dominant and lifts exact generation from near zero to near ceiling (Figure~\ref{fig:lever}; Figure~\ref{fig:p10all_final}). By 50M tokens $P(\mathrm{copy})$ has fallen from $.82$ to near zero and exact generation has risen to $.56$; by the end of training exact generation is near ceiling, whereas plain continuation stays copy-dominant.

\begin{figure}[t!]
\centering
\includegraphics[width=0.9\columnwidth]{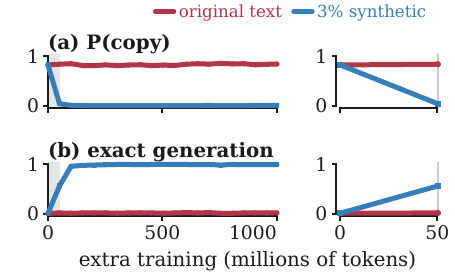}
\caption{Continued pretraining of SmolLM2-1.7B. The right panels zoom on the first 50M tokens. A $3\,\%$ synthetic mixture reverses copying by the first checkpoint, whereas mined passages that merely contain definitions do not (Figure~\ref{fig:p10all_final}).}
\label{fig:lever}
\end{figure}

The reversal costs nothing measurable on general text: held-out losses move by less than $.001$ at both scales while $D$ moves by more than twenty nats.

\textbf{Held-out versus all-shown.} The SmolLM2-360M experiments separate query-resolution practice from rehearsal of the missing cell. Two synthetic variants share format and data order. In the held-out condition each document queries combinations it does not show. In the all-shown condition every queried combination already appears on another stem in the same document, so the model never practices an absent cell. Both synthetic conditions turn to $D<0$ by the first checkpoint, after 50M tokens, and write the missing form on most prompts by the end of training, while plain continuation stays copy-dominant (Figure~\ref{fig:p7b_final}). The all-shown condition matches the held-out condition, so learning to resolve queries through temporary mappings does not require practicing an absent cell.

Full fine-tuning of Pythia-1.4B on prompts that show every requested form reproduces the all-shown result (Appendix~\ref{app:induce}), so the shift holds under full fine-tuning as well as continued pretraining.

\begin{figure}[t]
\centering
\includegraphics[width=0.95\columnwidth]{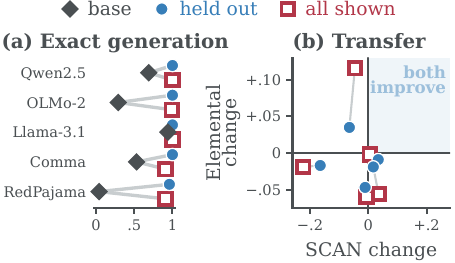}
\caption{LoRA takes exact generation to or near ceiling on five bases, while the external SCAN--Elemental pairs stay outside the shaded quadrant where both would improve: the gain stays with the practiced form.}
\label{fig:lora_boundary}
\end{figure}

Low-rank adapters trained on the same held-out or all-shown documents take five released 7--8B bases to or near the ceiling of the diagnostic (Figure~\ref{fig:lora_boundary}): RedPajama-7B, the one copy-dominant base, turns query-dominant, and the four that already composed deepen their query dependence. The gain stays with the practiced form: the external composition tasks register no gain under the pre-registered rule, SCAN-ICL falls on OLMo-2 and Llama-3.1, and OLMo-2's general-text loss passes the pre-registered bound (Appendix~\ref{app:part8}). The one ElementalTask gain arrives with a SCAN loss (Figure~\ref{fig:lora_boundary}b).

Transfer follows what was practiced, not composition tasks in general. At 1.7B the five external evaluations fall outside the pre-registered qualification rule, so their synthetic-versus-plain differences are reported as descriptive (Appendix~\ref{app:part10}). The boundary of transfer is therefore set by the 7--8B adapters, whose bases score well above floor on SCAN-ICL and ElementalTask-ICL and whose gains stay with the practiced form under that rule (Appendix~\ref{app:part8}).

\subsection{Transfer depends on the supervision}

\textbf{Definition exercises.} Definition exercises teach a different use of the prompt. A language model rewrote real passages into short exercises that define a label, term, or rule and then pose a query answerable only from that definition (verified examples in Table~\ref{tab:doc_examples}). Unlike passive exposure to definitions in natural text, every exercise requires an answer determined by a local definition. Fine-tuning SmolLM2-1.7B on 57M tokens of exercises is compared with the same schedule on the original documents, and both are evaluated on in-context classification whose class names are replaced by made-up symbols (made-up labels) or exchanged with one another (exchanged labels) \citep{wei2023symbol}, so that accuracy requires following the prompt's assignment. Made-up-label accuracy rises by $17.6$ points on TREC, $17.2$ on DBpedia, and $3.7$ on Yahoo (Figure~\ref{fig:overview}c), and exchanged-label accuracy rises on all three. On SmolLM3-3B the gains hold on TREC and Yahoo, and no counted evaluation loses. The exercises contain neither these datasets nor their labels. The transfer is specific to the practiced use of the prompt: a relational-reasoning suite that also draws definitions from the prompt (bAbI-1, bAbI-15, or CLUTRR) moves down by a few points and farther tasks move in both directions, so the exercises teach following prompt-assigned labels rather than prompt-dependent tasks in general. (Appendix~\ref{app:transfer}; Figure~\ref{fig:exercise_gains}).

The morphology diagnostic is untouched by the exercises: $P(\mathrm{copy})$ is unchanged and four-choice accuracy and exact generation stay near floor (Appendix~\ref{app:transfer}). The exercises make label following reusable across datasets and leave the morphology task in place.

\paragraph{The interventions separate suppression, construction, and transfer.}
Suppression moves the margin without producing the missing form. Training on the synthetic documents turns $D$ negative and lifts exact generation across model families, update methods, and encodings, and its transfer follows the practiced form. Definition exercises improve prompt-defined label use elsewhere while leaving the diagnostic unchanged. What mattered was which use of the prompt the training text repeatedly rewarded.

%% file: sections/07_discussion.tex
\section{Discussion}
\label{sec:discussion}

\begin{figure}[t]
\centering
\includegraphics[width=0.75\columnwidth]{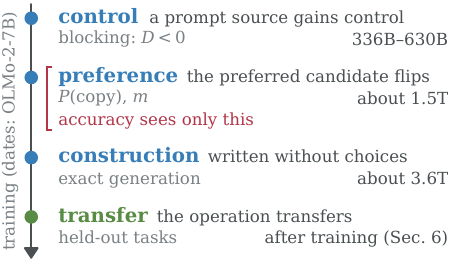}
\caption{Four events that accuracy compresses into one, in training order: a prompt source gains control of the answer, the preferred candidate flips (the only event accuracy registers), the form is written without choices, and the learned skill transfers to held-out tasks (after the training of Section~\ref{sec:lever}).}
\label{fig:events}
\end{figure}
The results separate four events that accuracy compresses into one (Figure~\ref{fig:events}). The requested value takes over control from the closest example before accuracy leaves chance.

Copying and composing from the requested value coexist in the same model. High-copy models already use the query to produce a shown combination, direct fine-tuning strengthens the effect of the requested value without erasing the effect of the closest example, and penalizing the copy-favoring heads moves the margin, while the missing form appears only once training supplies a way to build it.

\textbf{What the all-shown control shows.} In the all-shown control, documents in which every queried combination is already shown on another stem cross at the same checkpoint and reach comparable exact generation at 360M. The tie places the cause in what both conditions share, repeated resolution of queries through temporary mappings, and rules out rehearsal of the absent cell (Appendix~\ref{app:part7b}).

Developmental checkpoints should be read accordingly: a wrong answer can mark a substantive transition if its score now depends on the query rather than the closest example, and a correct multiple-choice answer need not establish construction if the form cannot be generated without candidates.

%% file: sections/07_conclusion.tex
\section{Conclusion}
\label{sec:conclusion}
On a diagnostic that separates copying from composition, in-context learning develops as control shifts among competing prompt cues. The requested value takes control of candidate scores before the correct answer wins. Suppressing copying moves the margin, practice at resolving queries writes the missing form, and training transfers the use of the prompt it repeatedly requires. Tracking control, preference, construction, and transfer shows what pretraining teaches beyond what accuracy reveals. Across models, checkpoint series, and training experiments, these four aspects develop on different schedules, so in-context learning is acquired as several capabilities, each on its own timetable.

%% file: sections/08_limitations.tex
\section*{Limitations}

Our diagnostic is a single controlled task that separates copying from composition, and the code and Quechua versions carry the contrast into two further encodings. In-context learning covers many other behaviors, and the four measures used here give a way to ask whether they develop in the same order. The training experiments use models up to 8B parameters and synthetic documents that share the diagnostic's format, so the transfer they produce is the format-bound transfer the design predicts. Larger models and more varied training data are the next tests of this account. The selected attention heads carry part of the behavior, so the localization rests on the prompt-span interventions. The pre-registered training block, transfer evaluations, and analysis-plan record are in Appendices~\ref{app:part7b}, \ref{app:transfer}, and~\ref{app:disclosures}.

%% file: sections/08b_ethics.tex
\section*{Ethical Considerations}
The Southern Quechua paradigms used in the morphology version of the task come from published linguistic resources \citep{adelaar2004andes,kirov2018unimorph,batsuren2022unimorph}.

The mined documents come from public collections of web, scientific, mathematical, encyclopedia, forum, and legal text. We applied no additional screening for personal or offensive content beyond the source collections' own filtering, so these sources may contain such material.

AI assistants were used for writing code and editing text.

%% file: sections/09_appendix.tex
\section*{Contents of the Appendix}
\begin{itemize}[noitemsep, topsep=0pt, leftmargin=2.2em]
    \item[\ref{app:prereg}] Task construction and conventions
    \item[\ref{app:onsets}] Threshold crossings during training
    \item[\ref{app:controls}] Final scores and prompt-variant controls
    \item[\ref{app:tables}] Blocking, deletion, and scoring checks
    \item[\ref{app:traj}] Training trajectories of seventeen runs
    \item[\ref{app:instantiations}] Code and Quechua versions
    \item[\ref{app:das}] Activation patching
    \item[\ref{app:stage2}] Stage-2 OLMo-2-7B models
    \item[\ref{app:induce}] Fine-tuning a copying model
    \item[\ref{app:confirm}] Re-test on new items
    \item[\ref{app:mech}] Selected attention heads
    \item[\ref{app:part7b}] Continued pretraining with synthetic documents and attention penalties (360M)
    \item[\ref{app:part10}] Continued pretraining with synthetic documents and mined passages (1.7B)
    \item[\ref{app:transfer}] Layer sweep, head penalties, adapters, and exercises
    \item[\ref{app:compute}] Release
    \item[\ref{app:disclosures}] Changes to the analysis plans
\end{itemize}

\section{Task construction and conventions}
\label{app:prereg}

\paragraph{Generator.} Every prompt draws fresh stems, words, endings and held-out combination, so no fixed stem--ending mapping solves the task. A suffix-scheme stem concatenates one of 36 initial strings, one of 9 vowels and one of 22 final strings. No stem repeats within a prompt, and the queried stem never appears in the demonstrations. Endings are drawn without replacement from 30 strings (\emph{il, nak, or, um, esh, oth, ka, tu, ri, mo, lin, dor, sa, ne, po, ta, ku, zi, fen, gor, lu, wa, ex, an, ir, on, ul, ek, ip, ab}), and line order is shuffled. Each prompt draws one first-feature pair from \emph{small}/\emph{big}, \emph{old}/\emph{new}, \emph{red}/\emph{blue} and \emph{hot}/\emph{cold}, and one second-feature pair from \emph{one}/\emph{many}, \emph{here}/\emph{there} and \emph{mine}/\emph{yours}. The code scheme keeps the first-feature pairs, takes the second from \emph{circle}/\emph{square}, \emph{north}/\emph{south} or \emph{left}/\emph{right}, and uses 20 two-character symbols with the header ``Codebook:'' and the separator ``=>''.

Four fixed prompt sets serve all models and checkpoints, two (320 items) for the initial analyses and two for the re-test on new items (Appendix~\ref{app:confirm}).

\paragraph{A literal prompt.} One generated prompt, verbatim:
\begin{quote}\small\ttfamily In this language:\\ one glon = glon-lu\\ many pliv = pliv-dor\\ one fooz = fooz-lu\\ big many swoov = swoov-an-dor\\ many bouz = bouz-dor\\ small one kraix = kraix-ek-lu\\ small one plut = plut-ek-lu\\ big stoosh = stoosh-an\\ big one lask = lask-an-lu\\ big one twix = twix-an-lu\\ big crood = crood-an\\ small berb = berb-ek\\ small plaind = plaind-ek\\ big many doot = doot-an-dor\\ small many glev =\end{quote}

The four scored continuations, each preceded by a space, are \texttt{glev-ek-dor} (correct), \texttt{glev-ek-lu} (the copy, sharing the first ending \texttt{-ek}), \texttt{glev-an-dor} and \texttt{glev-an-lu}. These are the candidates for four-choice accuracy and the ``ranks first'' columns.

\paragraph{Vocabulary.} ``Correct'' is the rule-consistent candidate of Equation~\ref{eq:candidates} in Section~\ref{sec:task}, and ``copied'' is the candidate that repeats the closest example's second ending. A generation is labeled by the first word of its first line, split on hyphens. One ending counts as truncated and a form outside the four combinations as malformed. Anything else is correct, copied, or another shown combination. ``Closest ending'' and ``requested value'' name the blocked spans.

\paragraph{Intervention conventions.} Layers are decoder blocks counted from zero. Activation patching replaces a block's output at the position of the requested second-feature value while the prompt is processed, before generation. The model then takes the most probable token at each step for up to eight new tokens. A blocked span covers every tokenizer token that overlaps the named string or strings, and blocking the closest ending covers the second-ending string in both lines of the closest example.

\paragraph{Tokenization.} All 30 endings are separate tokens after a hyphen in Pythia and Gemma \citep{biderman2023pythia,gemma3}, most are in OLMo-2, and Qwen merges some hyphens with their endings \citep{qwen3,olmo2025olmo2}. An ending can span several tokens, so the four forms can differ in token count. We score the next ending with the correct preceding tokens supplied only where hyphen and ending stay separate (scoring checks in Appendix~\ref{app:tables}).
\section{Threshold crossings during training}
\label{app:onsets}
Table~\ref{tab:onsets} dates when each behavior of Section~\ref{sec:dynamics} crosses a fixed threshold. Rows use the original items unless marked as the code scheme. Times are crossings of a fitted curve, not first significant differences. We fit each mean trajectory over $\log_{10}$ tokens (training steps for Pythia) by least squares to a four-parameter logistic curve, with both asymptotes between 0 and 1, rate at least .05, and midpoint within two decades of the observed range. Intervals are the 2.5th to 97.5th percentiles of the finite crossings over 300 resamples of the items with replacement. A resample whose fitted asymptote stays below the threshold is censored.

\begin{table}[htbp]
\centering\small
\setlength{\tabcolsep}{1.5pt}\renewcommand{\arraystretch}{0.95}
\begin{tabular}{@{}p{4.45cm}r@{}}
\toprule
measure (threshold) & crossing [95\% interval] \\
\midrule
\multicolumn{2}{@{}l}{\emph{OLMo-2-7B}} \\
copy the stem ($\geq .85$) & 11B [10, 12] \\
first feature correct ($\geq .70$) & 76B [66, 88] \\
second feature correct ($\geq .70$) & 338B [254, 480] \\
shown combinations, exact generation ($\geq .50$) & 224B [143, 293] \\
\heldout{}, four-choice accuracy ($> .25$) & 1,464B [1,270, 1,675] \\
\heldout{}, exact generation ($\geq .15$) & 3,586B [3,286, 4,020] \\
\midrule
\multicolumn{2}{@{}l}{\emph{OLMo-2-13B}} \\
first feature correct & 34B$^{\ddagger}$ \\
second feature correct & 245B$^{\ddagger}$ \\
shown combinations, exact generation & 174B [130, 368] \\
\heldout{}, four-choice accuracy & 1,825B [1,480, 2,307] \\
\heldout{}, exact generation & 8.4T$^{\dagger}$ \\
\midrule
\multicolumn{2}{@{}l}{\emph{Pythia-1.4B (training steps)}} \\
copy the stem & 2,037 [2,012, 2,118] \\
first feature correct & 25.9k [20.1k, 31.9k] \\
shown combinations, exact generation & 58.7k [44k, 84k] \\
second feature correct & censored$^{\ast}$ \\
\heldout{}, four-choice accuracy & censored$^{\ast}$ \\
\heldout{}, exact generation & censored$^{\ast}$ \\
\midrule
\multicolumn{2}{@{}l}{\emph{OLMo-2-7B, code scheme}} \\
\heldout{}, four-choice accuracy & 143B [115, 178] \\
\heldout{}, exact generation & 131B [103, 169] \\
shown combinations, exact generation & 815B [603, 1,218] \\
\bottomrule
\end{tabular}
\caption{Fitted threshold crossings. ``Copy the stem'' is the share of generations that begin with the requested novel stem, and ``first/second feature correct'' is the share of single-feature probes whose ending matches its prompt definition. Thresholds follow the first block unless stated. $^{\dagger}$Extrapolated beyond 5T tokens, with 8\% of resamples censored. $^{\ddagger}$Intervals were not retained for these two fits. $^{\ast}$77--100\% of resamples are censored. The re-test items give a four-choice crossing of 1.79T for OLMo-2-7B (Appendix~\ref{app:confirm}).}
\label{tab:onsets}
\end{table}

\paragraph{Ordering.} In OLMo-2-7B every component behavior has crossed by 338B, before or during the 336B--630B interval in which $D$ changes sign, while the held-out form crosses at 1,464B for four-choice accuracy and 3,586B for exact generation. OLMo-2-13B shows the same order, although its exact-generation fit extrapolates beyond the observed run. Pythia-1.4B reaches three of the six thresholds. In the code scheme, held-out accuracy and generation cross at 143B and 131B, before shown-combination generation at 815B, so the component-before-selection order is established for the suffix scheme only. The 1.79T re-test crossing and the 1,464B crossing come from different items, and neither alters the 336B--630B blocking interval.
\section{Final scores and prompt-variant controls}
\label{app:controls}
\label{app:finals}
\begin{figure}[t]
\centering
\includegraphics[width=\columnwidth]{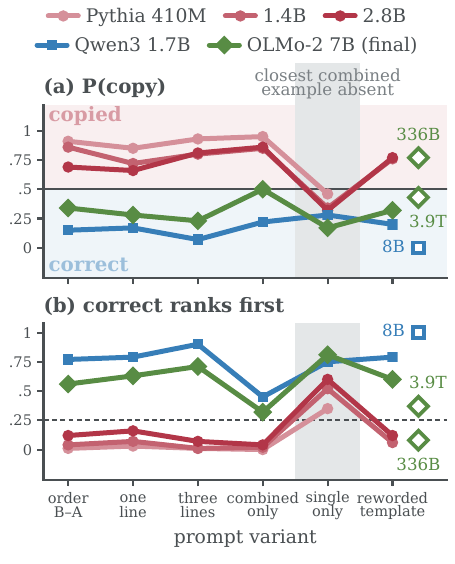}
\caption{Prompt-variant controls. (a) $P(\mathrm{copy})$ and (b) the fraction of prompts where the correct form ranks first. Series are Pythia-410M, Pythia-1.4B, Pythia-2.8B, Qwen3-1.7B, and released OLMo-2-7B. Qwen3-8B and the OLMo-2-7B 336B and 3.9T checkpoints appear only for the reworded template. Variants reverse ending order (B--A), use one or three example lines per combination, retain only combined-form or single-feature lines, or reword the template. The shaded single-feature-only condition has no closest combined example. Guides mark equal copied/correct probability ($.5$) and four-choice chance ($.25$).}
\label{fig:controls}
\end{figure}

Table~\ref{tab:fingerprint} lists the scores behind Figure~\ref{fig:fingerprint}. Five changes to the prompt's surface form test whether Pythia keeps preferring the copied answer and Qwen3-1.7B, Qwen3-8B and OLMo-2-7B the correct one (Figure~\ref{fig:controls}).

OLMo-2-1B ($.58$), Qwen3-0.6B ($.50$) and Qwen2.5-1.5B ($.49$) lie between the thresholds. Qwen2.5-0.5B ($.86$) is the only non-Pythia high-copy model. The correct form is generated on $.00$--$.05$ of prompts in the high-copy group and $.30$--$.93$ in the low-copy group (Table~\ref{tab:fingerprint}).

\InputIfFileExists{tables/fingerprint_app}{}{\InputIfFileExists{tables/fingerprint}{}{}}

\paragraph{Prompt variants.} Pythia's $P(\mathrm{copy})$ stays at $.66$--$.95$ under every variant keeping the closest example and falls to $.32$--$.46$ with single-feature lines only (Figure~\ref{fig:controls} and the supplement). Qwen3-1.7B and released OLMo-2-7B rank the correct form first on at least $.45$ of prompts under every variant but combined-form lines only. Qwen3-8B, run only on the reworded template, ranks it first on $1.00$.

\section{Blocking, deletion, and scoring checks}
\label{app:tables}
\label{app:deletion}
Table~\ref{tab:knockout} lists every blocked span per model in bfloat16 on the original items. The body and Figure~\ref{fig:knockout} use float32 re-test scores (Precision, below). Figure~\ref{fig:mech} shows one model per group.

\paragraph{Blocked spans.} Blocking the closest ending raises $m$ wherever measured, by $+1.00$ to $+5.34$ nats, and blocking the requested value lowers it by $-1.59$ to $-10.51$ (Table~\ref{tab:knockout}). Pythia's $D$ is positive ($+2.77$ to $+3.75$), OLMo-2-1B's is near zero ($-0.23$ [$-0.74,+0.29$]), and Qwen3-0.6B and all measured low-copy models have negative $D$ ($-1.36$ to $-9.44$). Single-feature lines affect every tested model and unrelated lines change little. Qwen3-4B's closest-ending block was not run.

\InputIfFileExists{tables/knockout_app}{}{\InputIfFileExists{tables/knockout_h}{}{}}

Blocking the closest ending moves Pythia-2.8B's $m$ from $-2.1$ to $+2.6$ nats and blocking the requested value moves Qwen3-8B's from $+8.9$ to $-1.5$, reversing the preferred answer in both (Figure~\ref{fig:mech} and Table~\ref{tab:confirm_finals}, re-test items in bfloat16).

\paragraph{Deleting lines.} Table~\ref{tab:deletion_final} repeats the test with lines deleted instead of masked. Deleting the closest example makes the correct answer preferred in Pythia-1.4B, Pythia-2.8B and Qwen3-0.6B ($+3.20$, $+3.30$ and $+3.52$ nats). The gain shrinks as unblocked $m$ rises, reaching $-0.19$ in Qwen3-8B. Deleting the correct-ending lines lowers $m$ everywhere ($-0.53$ to $-1.52$). Deleting unrelated lines changes $m$ by $-0.13$ and $+0.05$ in Pythia and by $-0.44$ to $-0.89$ in tested low-copy models. Eight of the twelve models in the blocking panel were tested (Table~\ref{tab:deletion_final}).

\input{tables/deletion_final_app}

\paragraph{Qwen2.5-0.5B.} This model was added after the model set was fixed and run on independently generated items. It is not part of the directional test. Blocking the closest ending raises $m$ by $+3.25$ nats and lowers $P(\mathrm{copy})$ from $.88$ to $.42$, blocking the requested value lowers $m$ by $.66$, and the other question value and unrelated lines move it by $+.6$ and $+.1$. Its bfloat16 $D$ is $+2.59$ [$+2.34,+2.85$], positive on $88\%$ of items, and float32 agrees to $.02$ nats. The activation patch of Figure~\ref{fig:patch} produces none of the 109 never-shown targets and a shown combination on $.36$ and $.31$ in the two item sets.

\paragraph{Scoring variants.} Restricting to single-token endings, scoring only the first token where the forms differ, or length-normalizing leaves $P(\mathrm{copy})$ within $.01$ and changes neither the sign of $D$ nor any model's group. Item-level first-token and summed scores correlate at $.84$--$1.0$ for $m$ and $.90$--$.98$ for $D$. All three put OLMo-2-7B's last positive $D$ at 336B and first negative at 630B, and OLMo-2-13B's first sign change at 168B--336B. Length normalization moves the correct-ranks-first fraction by at most $.02$ in final models and at most $.05$ across checkpoints (at 5B, where both scores are near chance). Two other checkpoints move by $.03$. A later analysis that conditions $D$ on full-prompt $m$ moves the OLMo-2-7B sign change to 168B--336B without replacing the unconditioned series (Appendix~\ref{app:disclosures}, supplement table).

\paragraph{Precision.} We re-scored $P(\mathrm{copy})$, the blocked-span changes in $m$, and $D$ in float32 on the re-test items and the OLMo-2-7B and OLMo-2-13B checkpoints, which the body and Figure~\ref{fig:knockout} use. In OLMo, $P(\mathrm{copy})$ changes by at most $.005$ and $D$ by at most $.25$ nats, pooled signs are unchanged, and item-level signs agree on $94$--$100\%$. In Pythia, bfloat16 overstates closest-ending changes by $13$--$62\%$ and $D$ by $12$--$105\%$ across models and item sets. Pooled float32 $D$ for Pythia-1.4B, -2.8B and -12B is $+2.4$, $+2.1$ and $+1.9$, against $+3.3$, $+2.4$ and $+3.7$ in bfloat16. Signs, groups and checkpoint ordering are unchanged. On 50 Pythia-1.4B items, padded batches and single sequences differ by under $.001$ nats in float32 and up to $2.6$ nats per form in bfloat16 (supplement table).
\section{Training trajectories of seventeen runs}
\label{app:traj}
\label{app:part7a}
\begin{figure*}[t]
\centering
\includegraphics[width=\textwidth]{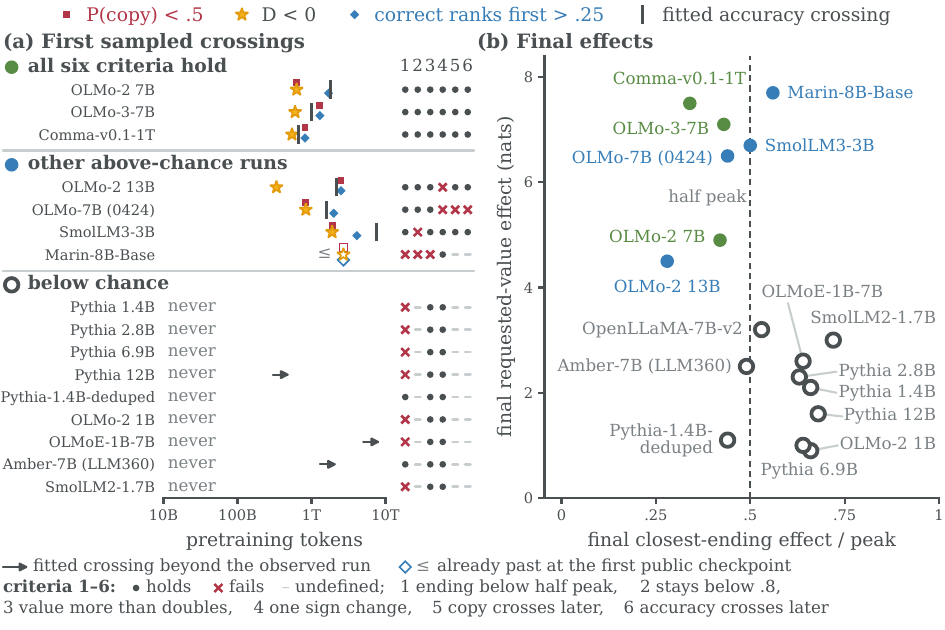}
\caption{Crossing dates and endpoints of the training trajectories. (a) First sampled crossings of $P(\mathrm{copy})<.5$, $D<0$, and correct-ranks-first $>.25$, with fitted accuracy crossings shown separately. ``Never marks an unobserved sampled crossing. Arrows mark fitted crossings beyond the observed run, and Marin is already above threshold at its first public checkpoint. The numbered checks are a closest-ending effect below half its peak, an effect that stays below $.8$ after $D<0$, a requested-value effect that more than doubles, at most one sign change, a copied-answer probability crossing after the last positive $D$, and a fitted accuracy crossing after the sign change. OpenLLaMA-7B-v2 is measured in optimizer steps: its first negative $D$ occurs at 460k, but it never crosses the probability or accuracy thresholds. (b) Final closest-ending effect divided by its peak versus final requested-value effect in nats. The groups distinguish runs meeting all ordering criteria, other above-chance runs, and below-chance runs. The vertical guide marks half the peak. Exact sampled and fitted dates are in Table~\ref{tab:crossings}.}
\label{fig:crossings}
\end{figure*}

The body's pretraining evidence comes from one lab (Ai2). Here we follow the eleven older public series and six models selected from a screen of sixteen other released models (Table~\ref{tab:p7a_stagea}), asking in each whether $D$ turns negative before four-choice accuracy rises. Figure~\ref{fig:crossings} and Table~\ref{tab:crossings} date the crossings.

\paragraph{Runs and screen.} Seven trajectories end above four-choice chance: OLMo-2-7B, OLMo-2-13B, OLMo-7B-0424, OLMo-3-7B, Marin-8B-Base, SmolLM3-3B and Comma-v0.1-1T. The other ten end below chance. The screen keeps models whose correct form ranks first on at least 160 of 320 items. We follow all four, plus OpenLLaMA-7B-v2 ($.17$) and SmolLM2-1.7B ($.12$), the two low-accuracy models with the highest correct-ranks-first rates. Apertus-8B-2509 ($.42$) clears only the table's partial threshold. OLMo-3 is another Ai2 run, so it is not independent evidence.

\InputIfFileExists{tables/p7a_stagea_app}{}{\InputIfFileExists{tables/p7a_stagea}{}{}}

\paragraph{Crossings.} Figure~\ref{fig:crossings}a dates the first sampled crossing of each threshold. In every run ending above chance, the first negative $D$ comes before or at the sampled accuracy crossing, though a first crossing need not last. No below-chance run reaches a sustained negative $D$.

\InputIfFileExists{tables/crossings_app}{}{\InputIfFileExists{tables/trajdesc_crossings_h}{}{}}

\paragraph{Ordering criteria and exceptions.} OLMo-2-7B, OLMo-3-7B and Comma-v0.1-1T meet all six checks in the figure caption. Their sign-change brackets are 336B--630B, 411B--600B and 482B--545B. OLMo-2-13B meets five but changes sign three times, first at 168B--336B, back above zero at 630B, and again at 630B--839B. SmolLM3-3B shifts before accuracy rises, but its closest-ending effect later climbs back above $.8$ of its peak. Marin-8B-Base is already past both thresholds at its first public checkpoint, so its ordering cannot be dated. OpenLLaMA-7B-v2's mean $D$ dips to $-0.20$ only at its last checkpoint, and SmolLM2-1.7B's stays positive.

OLMo-7B-0424 is the exception. Its first negative $D$ is at 838B, but it hovers around zero until 2.3T, while fitted accuracy crosses at 1.6T and sampled accuracy exceeds $.25$ at 2.0T. The run fails three of the six checks.

After 1T in OLMo-2-7B, the requested-value effect moves between $2.9$ and $5.3$ nats across neighbouring checkpoints, and the closest-ending effect briefly reaches $3.1$ at 1.68T. The two never re-cross, and $D$ stays negative at every later checkpoint (Figure~\ref{fig:trajcol}).

\paragraph{Endpoints.} The seven above-chance runs finish with requested-value effects of $4.5$--$7.7$ nats and closest-ending effects at $.28$--$.56$ of their observed peaks. The ten below-chance runs finish at $.9$--$3.2$ nats and $.44$--$.72$ of peak (Figure~\ref{fig:crossings}b). Pythia-1.4B-deduped and Amber-7B do fall below half their peaks ($.44$ and $.49$), yet keep positive $D$ ($+3.1$, $+1.6$) and high $P(\mathrm{copy})$ ($.80$, $.72$). The released OLMo-2-1B lies outside its stage-1 trajectory, at $P(\mathrm{copy})=.57$ and $D=-0.6$ [$-1.0,-0.2$]. Appendix~\ref{app:stage2} covers the same stage-1 versus released difference for OLMo-2-7B.
\section{Code and Quechua versions}
\label{app:instantiations}
\begin{figure}[t]
\centering
\includegraphics[width=\columnwidth]{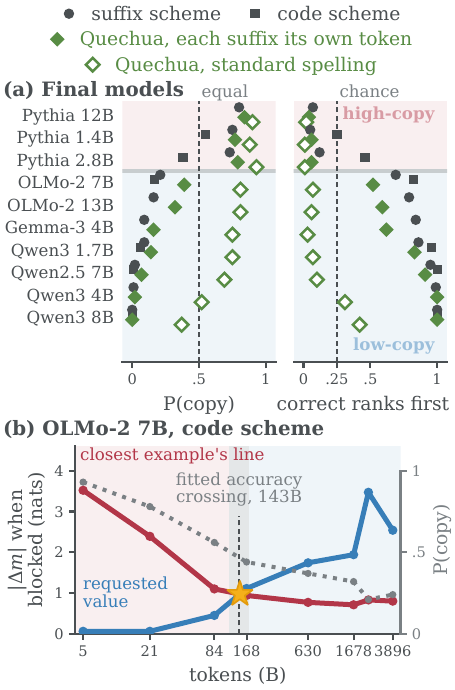}
\caption{Surface-form and tokenization checks. (a) $P(\mathrm{copy})$ and correct-ranks-first for the suffix scheme, code scheme, segmented Quechua, and Quechua in standard spelling. Six models have code results and ten have Quechua results. Guides mark $.5$ and $.25$. The broad model ordering survives the code and segmented-Quechua versions, while standard spelling reduces accuracy in every model. (b) OLMo-2-7B in the code scheme: magnitudes of the changes in $m$ from blocking the closest example's line or the requested value, with $P(\mathrm{copy})$ on the other axis. The blocking turn at 168B overlaps the fitted accuracy crossing near 143B (Table~\ref{tab:onsets}).}
\label{fig:generality}
\end{figure}

Two further versions test whether the model ordering of Section~\ref{sec:fingerprint} depends on the invented suffixes or on tokenization (Figure~\ref{fig:generality}). One uses code words and symbols in place of endings. The other uses Southern Quechua noun paradigms, scored with each suffix as its own token and in standard spelling.

\paragraph{Code scheme.} The six tested models keep the suffix-scheme ordering and score better. $P(\mathrm{copy})$ is $.55$, $.38$, $.17$, $.06$, $.01$ and $.00$ for Pythia-1.4B, Pythia-2.8B, OLMo-2-7B, Qwen3-1.7B, Qwen2.5-7B and Qwen3-8B (Figure~\ref{fig:generality}a). Blocking here covers the closest example's whole line, so its size is not comparable with suffix-ending blocking. In OLMo-2-7B the requested-value effect first exceeds the closest-example effect at 168B, inside the interval of the fitted four-choice crossing, 143B [115, 178]. The two events coincide at this resolution, whereas in the suffix scheme the blocking turn came first (Figure~\ref{fig:generality}b).

\paragraph{Quechua construction.} Southern Quechua nouns combine possessor and case endings \citep{adelaar2004andes,kirov2018unimorph,batsuren2022unimorph}. Of the 494 UniMorph \texttt{que} noun lemmas we keep the 242 vowel-final, lower-case ASCII lemmas whose cells are exact concatenations of lemma, possessor and case (consonant-final lemmas would need epenthesis), minus function words, loans and proper names. Possessor endings are \emph{-y, -yki, -n, -nchik, -nku}, and case endings \emph{-man, -manta, -pi, -wan, -kama, -paq, -rayku}. Excluding prefix-related pairs (\emph{y}/\emph{yki}, \emph{n}/\emph{nchik}, \emph{n}/\emph{nku}, \emph{man}/\emph{manta}) leaves 7 possessor pairs and 20 case pairs. Each prompt draws one pair from each pool, and the closest example shares the possessor. Segmented forms look like \texttt{tanta-y-pi}, standard spelling like \texttt{tantaypi}.

\paragraph{Tokenization and scores.} With suffixes tokenized separately, Pythia has $P(\mathrm{copy})=.77$--$.84$ and correct-ranks-first $.04$--$.06$. Qwen3-1.7B, Qwen2.5-7B, Qwen3-4B and Qwen3-8B have $P(\mathrm{copy})=.00$--$.14$ and correct-ranks-first $.83$--$1.00$, and OLMo-2 lies between. Blocking the requested value lowers $m$ by $-1.4$ to $-8.5$ nats, while blocking the closest ending raises it by $+.8$ to $+5.3$. Standard spelling, where the tokenizer often fuses the stem-final vowel with the possessor suffix, drops eight models to correct-ranks-first $.01$--$.10$ (Qwen3-4B $.31$, Qwen3-8B $.42$) and raises $P(\mathrm{copy})$ to $.37$--$.93$, though requested-value blocking still orders the models ($-1.1$ to $-6.0$ nats). Models do not consistently prefer the copy, which shares the possessor, over the shown form that shares the case, so Quechua shows a pull toward shown forms as such.

\paragraph{Contamination check.} Four-choice accuracy is $.21$--$.31$ with only the Quechua header and query, and $.22$--$.29$ with demonstrations whose possessors and cases differ from the query's. All Wilson intervals contain $.25$ except OLMo-2-13B's header-only $.31$ [$.26,.37$], so we show that model's Quechua scores but exclude them from the ordering. No model generates a correct form on any of the 320 header-only items. Training results on the code and Quechua versions are in Appendix~\ref{app:part10}, adapter results in Appendix~\ref{app:transfer}.
\section{Activation patching}
\label{app:das}
\begin{figure}[htbp]
\centering
\includegraphics[width=\columnwidth]{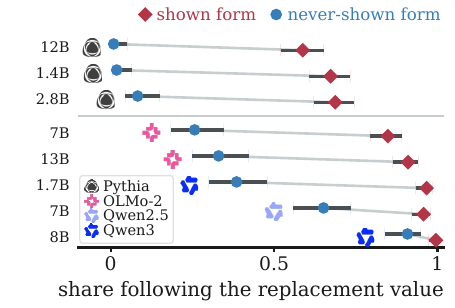}
\caption{Activation patching after layer 4 at the requested-value word. The two series give the shares producing the complete form named by the replacement value when that target is never shown or already shown. Models are grouped into high-copy and low-copy and ordered by their never-shown share. Both conditions start from a shown base form.}
\label{fig:patch}
\end{figure}

Under the patch of Section~\ref{sec:task}, never-shown shares in the low-copy group rise as $P(\mathrm{copy})$ falls: $.26$ for OLMo-2-7B, $.33$ for OLMo-2-13B, $.39$ for Qwen3-1.7B, $.65$ for Qwen2.5-7B and $.91$ for Qwen3-8B. Shown shares are $.85$--$1.00$. Pythia's shown share is $.59$--$.69$, but its never-shown share never exceeds $.08$ (Figure~\ref{fig:patch}). Counting only the requested ending, regardless of stem or other ending, Pythia's never-shown shares rise to $.11$, $.31$ and $.19$ while the low-copy models move by at most $.03$ (supplement table). High-copy models can produce the requested ending without building the full form.

\paragraph{Why this position and layer?} Distributed alignment search \citep{wu2023interpretability} on the final Pythia-1.4B fits an 8-dimensional rotated subspace at the requested-value word, the last stem token and ``=''. Only the requested-value word responds, most strongly after layer 4, where the greedy ending changes on most items (an approximation) for the first feature and about half for the second. Transfer is $.97$ on the three fitted word pairs but $.04$ on the held-out pair. The layer-4 success rate rises from $.28$ at step 2k (about 4B tokens) to $.95$ at step 143k (300B). Layer 4, the Pythia-1.4B optimum, is used as the shared early layer for every model, although per-model sweeps peak at layers 0, 1, 9, 11 and 14 (Appendix~\ref{app:transfer}).

Both conditions in Figure~\ref{fig:patch} start from a shown base form, and the never-shown base-form condition was not run. Position-by-position patching retains most of the change in $m$ through the first $.33$ of layers in Pythia-1.4B and $.53$--$.64$ in low-copy models. This depth measurement replaces a planned between-group comparison at the requested-value word that could not be evaluated as specified (Appendix~\ref{app:disclosures}).
\section{Stage-2 OLMo-2-7B models}
\label{app:stage2}
The released OLMo-2-7B in Table~\ref{tab:fingerprint} ranks the correct form first on .69 of prompts and writes it on .30, above the .53 and .17 at 3.9T of the last stage-1 checkpoint, which ends the trajectory of Figure~\ref{fig:trajcol}. The release is a parameter average of three stage-2 runs on different data mixtures \citep{olmo2025olmo2}. In the one run with public intermediate checkpoints, correct-ranks-first rises from $.53$ to $.64$ and exact generation from $.17$ to $.22$ within 5B additional tokens. Over 9B--50B, correct-ranks-first stays mostly at $.63$--$.68$, dipping to $.55$ at 17B, and exact generation ranges over $.15$--$.27$. A gain is therefore already present during stage-2 training, before averaging. The comparison cannot separate further training from the change of data mixture.

\InputIfFileExists{tables/stage2_app}{}{\input{tables_supplement/stage2}}
\section{Fine-tuning a copying model}
\label{app:induce}
We ask whether direct fine-tuning of Pythia-1.4B reproduces the pretraining order, $D$ changing sign before accuracy rises, and whether the all-shown result of Section~\ref{sec:lever} also holds under fine-tuning.

\paragraph{Design.} We fully fine-tune released Pythia-1.4B for 2,000 steps (learning rate $10^{-5}$, warm-up, cosine decay, batch size 32). Training uses its own prompt template, evaluation the standard one. Two seeds target the correct held-out form. A third run queries only shown combinations, mirroring the all-shown condition of Section~\ref{sec:lever}. We predicted that the bootstrap interval for $D$ would fall below zero at least 250 steps before the Wilson lower bound for four-choice accuracy exceeded $.25$, in both target runs, and that neither event would occur in the all-shown run.

\paragraph{Outcome.} Both target runs cross both thresholds at the first evaluation, step 250 (supplement figure). On a 10-step grid, accuracy clears chance at step 10 or 20 and passes $.9$ by step 50, while $D$ turns negative at step 20 in both runs. The closest-ending effect falls from $5.3$ to about $3$ nats and then fluctuates between $+2.3$ and $+7.6$, while the requested-value effect grows from $2$ to more than $20$. The all-shown run also learns the held-out form. At step 250 it has $P(\mathrm{copy})=.83$, a closest-ending effect near $31$ nats and $D=+21$. By step 500, $D$ is negative and correct-ranks-first is $.50$, rising to $.70$--$.78$ from step 1,000 with closest-ending and requested-value effects near $10$ and $30$ nats. Neither part of the predicted pretraining order is reproduced. The requested-value effect grows while the closest-ending effect is still present, instead of after it disappears.
\section{Re-test on new items}
\label{app:confirm}
We repeat the tests of Sections~\ref{sec:fingerprint} and~\ref{sec:dynamics} on 320 new prompts per model, generated after those analyses were fixed (Table~\ref{tab:confirm_finals}). The re-test covers new prompts from the same generator, not independently trained models.

\paragraph{Criteria.} $P(\mathrm{copy})$ must be at least $.70$ in the high-copy group and at most $.30$ in the low-copy group, $D$ must take the predicted sign in each group, and the unrelated-line 95\% interval must lie within $\pm .3$ nats. OLMo-2-7B's sign change must fall in 336B--630B and its fitted four-choice accuracy must cross chance afterwards. OLMo-2-1B and Qwen3-0.6B have no directional prediction, leaving nine in the sign test.

\paragraph{Results.} All probability thresholds and all nine signs hold (Table~\ref{tab:confirm_finals}). The 336B--630B interval recurs in $.999$ of joint item resamples, and the 1.79T accuracy crossing follows the sign change in every resample. The unrelated-line interval fails for Pythia-1.4B and Pythia-12B in bfloat16, with means of only $+.24$ and $+.17$, and passes in the float32 re-scores (Appendix~\ref{app:tables}). The six trajectory outcomes of Appendix~\ref{app:traj} are unchanged. OLMo-2-13B still shows three sign changes, and the low-accuracy OLMo-2-1B, OLMoE and standard Pythia runs never turn negative.

\input{tables/confirm_finals_app}
\section{Selected attention heads}
\label{app:mech}
\label{app:heads}
\label{app:part5}

\begin{figure}[htbp]
\centering
\includegraphics[width=\columnwidth]{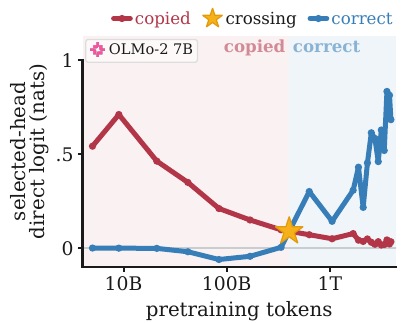}
\caption{Direct contributions toward each set's own answer, in nats, across OLMo-2-7B training, for the five heads selected for the copied answer and the five for the correct answer. Swapping selection and evaluation prompt sets moves the crossing to 1.68T. The star marks the crossing, inside the 336B--630B interval where whole-model $D$ changes sign.}
\label{fig:heads_traj}
\end{figure}

\paragraph{Head contributions across training.} Across OLMo-2-7B training (Figure~\ref{fig:heads_traj}), the copy-favoring set's direct contribution peaks at $.71$ nats at 9B, falls below $.10$ by 336B, and is about $.02$--$.05$ at most later checkpoints. The correct-favoring set stays near zero until 336B (dipping to $-.06$ at 84B--168B), reaches $.30$ at 630B, dips to $.14$ at 1049B, and exceeds $.6$ only after 2.5T ($.83$ at 3.57T). Heads are re-selected at each checkpoint. Swapping the selection and evaluation item sets moves the crossing from 630B to 1.68T (Figure~\ref{fig:heads_traj}), so this timing is less stable than the whole-model blocking interval.

\paragraph{Stability and contribution.} Selected twice, once on each item set, correct-answer heads overlap (Jaccard) at $.67$--$1$ in five of six low-copy models and $.43$ in OLMo-2-13B. Copy-answer heads overlap at $.67$ except in OLMo-2-13B ($.43$). Pythia overlaps are $.25$ to $.67$. In every low-copy model, direct logit attribution \citep{wang2023ioi} credits the selected heads with more of the correct-answer contribution than the multilayer-perceptron blocks, and with $.14$ to $.64$ of all head contributions toward that answer. Only Gemma-3-4B reaches half for both sets ($.64$ and $.56$), and no model's five-head sets explain half the matching token-blocking effect in both selection directions (Section~\ref{sec:heads}, Figure~\ref{fig:attention}b).

\paragraph{Silencing selected heads.} Figure~\ref{fig:attention}b averages the two directions. Silencing the copy-favoring sets raises $m$ by $.7$--$2.0$ nats in Pythia-1.4B, Qwen3-1.7B and the OLMo-2-7B checkpoints from 168B to 1049B, but only $.06$ in Qwen3-8B, within its random-set range. Silencing the correct-favoring sets lowers $m$ by $.4$--$1.2$ nats. The copy-head effect is as large at 168B as at 1049B, so these heads do not fall silent when control switches. The threshold is twice the 95th percentile of twenty layer-matched random sets in each direction. Copy-favoring sets clear it both ways in Qwen3-1.7B and OLMo-2-7B at 168B--1049B, but final OLMo-2-7B and Pythia-1.4B clear it one way only, which the averaged points hide. Correct-favoring sets clear it both ways in Pythia-1.4B and OLMo-2-7B and one way in the Qwen3 models.

\paragraph{Induction-head control.} A separate, task-independent control uses Pythia's 16 heads with the highest induction score on a random 40-token repeat \citep{olsson2022induction}. Replacing their answer-position outputs with those from other prompts raises correct-next-ending accuracy, given correct preceding tokens, from $.02$ to $.36$ in Pythia-1.4B at step 4k ($.39$ and $.33$ on the two item sets; random controls $.02$--$.03$) and from $.18$ to $.49$ in final Pythia-2.8B. Mean replacement gives $.44$ in Pythia-2.8B. In free generation, shown-combination accuracy falls from $.66$ to $.26$ and $.81$ to $.56$, so copying weakens, but held-out exact generation rises only from $.03$ to $.07$ and $.05$ to $.11$.

\input{sections/highlight/09b_part7b_h}
\input{sections/highlight/09e_part10_h}
\section{Layer sweep, head penalties, adapters, and exercises}
\label{app:transfer}
\label{app:part6}
\label{app:part8}
\label{app:part9}
\label{app:part10b}
Full tables are in the supplement. Figure~\ref{fig:exercise_gains} and Table~\ref{tab:doc_examples} cover the exercises.

\begin{figure*}[!t]
\centering
\includegraphics[width=\textwidth]{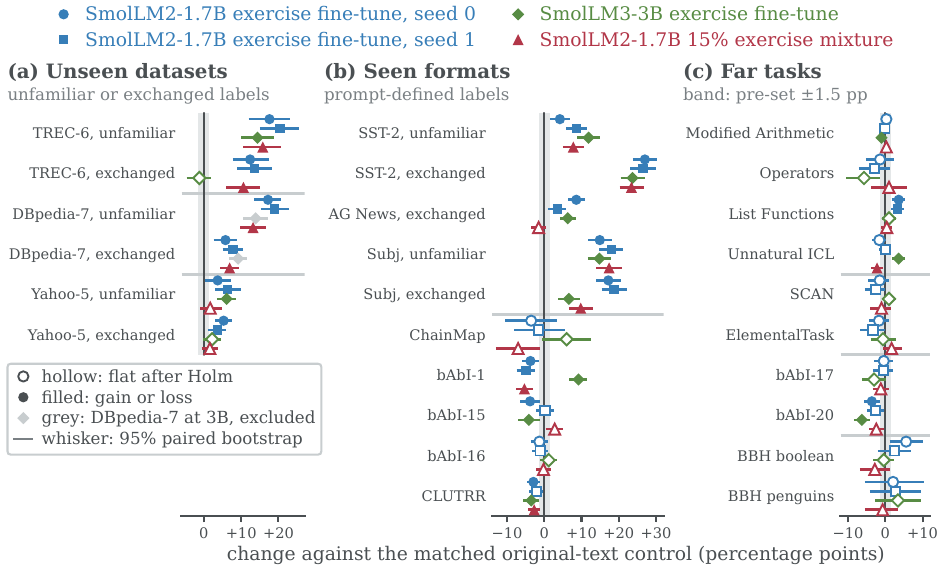}
\caption{Exercise-training changes against original-text controls, in percentage points. The series are two SmolLM2-1.7B exercise fine-tunes, a SmolLM3-3B exercise fine-tune, and a SmolLM2-1.7B 15\% exercise mixture. Rows separate made-up or exchanged labels on unseen datasets; prompt-defined labels and relations in seen formats; and farther tasks, including arithmetic, operators, List Functions, Unnatural In-Context Learning, SCAN, ElementalTask, bAbI and BIG-Bench Hard. Filled markers are significant after correction for multiple comparisons, hollow ones are not; the band marks the $\pm1.5$-point threshold. Grey DBpedia-7 results are excluded from 3B testing by the base-only qualification rule. The mixture control matches total tokens, not natural-text amount or repetition (Appendix~\ref{app:disclosures}).}
\label{fig:exercise_gains}
\end{figure*}

\InputIfFileExists{tables/doc_examples}{}{}

After Holm correction, a group of evaluations improves if some score rises and none falls, has a cost if any falls, and is otherwise unchanged. Controls use original text on the same schedules.

\paragraph{Layer sweep.} Activation-patching effects peak at layers 0/1 in Pythia-1.4B, 9/0 in OLMo-2-7B, 0/1 in Qwen3-1.7B, 1/1 in Qwen2.5-7B, and 14/11 in Gemma-3-4B (each pair selected on one item set and evaluated on the other). The body's layer 4 is a shared early layer from Pythia's alignment search, not a common peak (Appendix~\ref{app:das}).

\paragraph{Head outputs.} Neither multiplying selected-head outputs in thirteen models (against 19 layer-matched random head sets) nor fine-tuning Pythia-1.4B with head-derived penalties in five conditions beats its random-head or plain fine-tuning control. At SmolLM2-360M, an attention penalty cuts selected-head attention from $.310/ .341$ to $.008/ .006$ while $D$ stays positive at $+1.01/ +1.60$ (Appendix~\ref{app:part7b}). Transfer requires a gain of at least $.05$ with a paired interval above zero on both measures of both tasks in both runs, which lower attention alone does not give.

\paragraph{Adapters.} Low-rank adapters \citep{hu2022lora} train on 20M synthetic and 20M original-text tokens for five released 7--8B bases, with held-out or all-shown documents. Base exact generation is $.04$ (RedPajama), $.29$ (OLMo-2), $.53$ (Comma), $.69$ (Qwen2.5) and $.94$ (Llama-3.1). Adapters reach $.91$--$1.00$, yet none meets the joint transfer criterion on the three external tasks (Figure~\ref{fig:lora_boundary}). Only all-shown Llama-3.1 gains on ElementalTask ($+.12$), with a SCAN loss ($-.07$). OLMo-2 loses $.17$--$.22$ on SCAN and its Pile loss rises $+.035$ nats.

\paragraph{Attention.} Answer-position attention to the defining ending, relative to the closest ending, separates the model groups with area under the receiver-operating-characteristic curve $.814$ [$.59,1.00$], below the $.90$ threshold. None of the three attention predictions is met, and no trajectory shows the predicted increase. The measure is correlational.

\paragraph{Benchmarks and label flips.} Synthetic documents change none of ten standard benchmarks at 1.7B. At 360M, MMLU loses $.4$ points \citep{hendrycks2021mmlu}, LAMBADA gains $.6$ \citep{paperno2016lambada}, and the other eight are unchanged. With label meanings exchanged, 1.7B loses $3.2$ points on SST-2 \citep{socher2013sst} and $1.5$ on AG News \citep{zhang2015charcnn}. At 360M, SST-2 loses $2.1$, AG News gains $1.2$, and two others are unchanged.

\paragraph{Rewrites.} Every rewritten-document run stops at its first scheduled evaluation. The 1.7B run stops at 120 steps (250M tokens) when bAbI-20 falls $2.8$ points (Holm $p=.04$) \citep{weston2015babi}, and an extension to 360 steps stops again when List Functions falls $2.4$ points \citep{srivastava2023beyond}. Neither shows a full-window gain over original-document training.

\paragraph{Exercises.} Figure~\ref{fig:exercise_gains} shows exercise fine-tunes and a 15\% mixture against original-text controls. At 1.7B, made-up labels on unseen datasets gain $17.6$--$20.4$ points on TREC-6, $17.2$--$19.0$ on DBpedia-7, and $3.7$--$6.3$ on Yahoo-5 \citep{li2002trec,zhang2015charcnn}, and exchanged labels gain $3.6$--$13.6$. At 3B, TREC-6 and Yahoo-5 gain $14.4$ and $6.1$ points and no qualified unseen-label evaluation loses. Seen formats show gains (SST-2 exchanged labels $+23.4$, Subj $+17.4$ in the mixture) and losses (bAbI-1 $-5.3$, CLUTRR $-2.7$) \citep{weston2015babi,sinha2019clutrr}, as do farther tasks (List Functions $+3.3$--$+3.6$ at 1.7B, bAbI-20 $-3.6$ at 1.7B and $-6.3$ at 3B) \citep{srivastava2023beyond}. The morphology diagnostic is unchanged. The exercises practice prompt-defined labels \citep{wei2023symbol} (Table~\ref{tab:doc_examples}). Mined and filtered passages are in Appendix~\ref{app:part10}.
\section{Release}
\label{app:compute}
We release the code, per-item results for every table and figure, the analysis plans, and the full appendix as a supplement PDF (the `suppl.' tables cited here). Fine-tuned weights are regenerable from pinned base revisions and seeds and are not released. Experiments used about 320 GPU-hours on single H100 80GB GPUs.
\section{Changes to the analysis plans}
\label{app:disclosures}
Three choices made after the plans were fixed change a reported value or verdict. The blocking results are re-scored  (Appendix~\ref{app:tables}). A corrected Quechua scoring line moves the synthetic-document comparison from fail to pass (Appendix~\ref{app:part10}). The attention-matched control spans are split post hoc. No post hoc choice or descriptive result turns an unmet criterion into a met one. The supplement lists every departure.

%% file: tables/fingerprint_app.tex
\begin{table}[htbp]
\centering\setlength{\tabcolsep}{3pt}\renewcommand{\arraystretch}{0.95}
\setlength{\tabcolsep}{2.5pt}\scriptsize
\begin{tabular}{@{}lcccccc@{}}
\toprule
 & & & \multicolumn{4}{c}{generations} \\
\cmidrule(lr){4-7}
Model & $P(\mathrm{copy})$ & \shortstack{ranks\\first} & correct & copied & other & \shortstack{one\\ending} \\
\midrule
Pythia 410M & .91 & .03 & .01 & .13 & .30 & .56 \\
Qwen2.5 0.5B & .86 & .01 & .00 & .07 & .18 & .75 \\
Pythia 6.9B & .83 & .03 & .01 & .35 & .30 & .34 \\
Pythia 12B & .80 & .07 & .03 & .47 & .16 & .33 \\
Pythia 1.4B & .75 & .05 & .01 & .19 & .42 & .38 \\
Pythia 2.8B & .73 & .12 & .05 & .36 & .14 & .42 \\
OLMo-2 1B & .58 & .11 & .01 & .03 & .12 & .83 \\
Qwen3 0.6B & .50 & .34 & .06 & .05 & .03 & .86 \\
Qwen2.5 1.5B & .49 & .17 & .04 & .03 & .17 & .76 \\
\midrule
OLMo-2 7B & .21 & .69 & .30 & .04 & .03 & .63 \\
OLMo-2 13B & .16 & .79 & .33 & .02 & .01 & .62 \\
Qwen2.5 3B & .10 & .83 & .57 & .02 & .05 & .35 \\
Gemma-3 4B & .09 & .84 & .54 & .02 & .06 & .34 \\
Qwen3 1.7B & .09 & .86 & .43 & .01 & .03 & .52 \\
Qwen2.5 7B & .02 & .95 & .72 & .00 & .02 & .25 \\
Qwen3 4B & .01 & .99 & .80 & .00 & .00 & .17 \\
Qwen3 8B & .00 & 1.00 & .93 & .00 & .00 & .04 \\
\bottomrule
\end{tabular}
\caption{Final-checkpoint suffix-task scores (two item sets, 320 items). $P(\mathrm{copy})=p_{\mathrm{copied}}/(p_{\mathrm{copied}}+p_{\mathrm{correct}})$; $.5$ means equal probability. ``Ranks first'' is the share of prompts where the correct form leads four candidates (chance $.25$) under summed log-probabilities; the length-normalized share differs by at most $.02$ for every model. Generation columns omit malformed outputs. OLMo-2 7B is the released stage-2 parameter average (Appendix~\ref{app:stage2}). Qwen2.5-Coder-1.5B ($P(\mathrm{copy})=.42$) and Coder-7B ($.02$) are omitted.}
\label{tab:fingerprint}
\end{table}

%% file: tables/knockout_app.tex
\begin{table*}[htbp]
\centering\setlength{\tabcolsep}{2pt}\renewcommand{\arraystretch}{0.95}
\setlength{\tabcolsep}{2.5pt}\scriptsize
\begin{tabular}{@{}lrr rr rr rrr l@{}}
\toprule
 & & & \multicolumn{2}{c}{\textbf{closest example}} & \multicolumn{2}{c}{\textbf{question words}} & \multicolumn{3}{c}{\textbf{controls}} & \\
\cmidrule(lr){4-5}\cmidrule(lr){6-7}\cmidrule(lr){8-10}
Model & $P(\mathrm{copy})$ & \shortstack{$m$\\(unblocked)} & ending & \shortstack{two\\lines} & asked & other & unrelated & \shortstack{single-feature\\lines} & \shortstack{their\\endings} & \multicolumn{1}{c}{$D$ [95\% interval]} \\
\midrule
Qwen2.5 0.5B$^{\ddagger}$ & .88 & -- & $+3.25$ & -- & $-0.66$ & -- & -- & -- & -- & $+2.59$ $[+2.34,+2.85]$ \\
Pythia 12B & .84 & $-3.22$ & $+5.34$ & $+4.87$ & $-1.59$ & $+1.81$ & $-0.00$ & $-1.99$ & $-2.38$ & $+3.75$ $[+3.37,+4.09]$ \\
Pythia 1.4B & .81 & $-2.71$ & $+5.34$ & $+4.72$ & $-1.90$ & $+1.26$ & $+0.13$ & $-1.90$ & $-2.33$ & $+3.44$ $[+3.05,+3.82]$ \\
Pythia 2.8B & .79 & $-2.15$ & $+4.72$ & $+4.28$ & $-1.95$ & $+1.13$ & $+0.18$ & $-1.99$ & $-2.60$ & $+2.77$ $[+2.50,+3.04]$ \\
\midrule
OLMo-2 1B & .62 & $-0.58$ & $+2.94$$^{\dagger}$ & $+3.59$ & $-3.18$ & $+0.89$$^{\dagger}$ & $+0.17$ & $-1.67$ & $-1.93$$^{\dagger}$ & $-0.23$ $[-0.74,+0.29]$ \\
Qwen3 0.6B & .54 & $-0.18$ & $+2.94$ & $+3.70$ & $-4.30$ & $+0.85$ & $+0.01$ & $-1.33$ & $-1.78$ & $-1.36$ $[-1.68,-1.04]$ \\
\midrule
OLMo-2 7B & .24 & $+2.31$ & $+2.04$ & $+2.35$ & $-5.33$ & $-0.71$ & $+0.06$ & $-1.79$ & $-2.04$ & $-3.29$ $[-3.72,-2.89]$ \\
OLMo-2 13B & .17 & $+3.55$ & $+1.58$ & $+2.04$ & $-6.91$ & $-1.59$ & $-0.14$ & $-2.07$ & $-2.44$ & $-5.33$ $[-5.89,-4.78]$ \\
Gemma-3 4B & .11 & $+3.72$ & $+2.95$ & $+2.58$ & $-6.88$ & $-1.27$ & $-0.09$ & $-1.65$ & $-2.12$ & $-3.93$ $[-4.30,-3.54]$ \\
Qwen3 1.7B & .11 & $+3.84$ & $+1.86$ & $+2.18$ & $-7.02$ & $-0.08$ & $-0.11$ & $-1.58$ & $-1.92$ & $-5.15$ $[-5.62,-4.73]$ \\
Qwen2.5 7B & .02 & $+5.44$ & $+1.00$ & $+1.21$ & $-7.84$ & $-1.68$ & $+0.01$ & $-1.19$ & $-1.29$ & $-6.83$ $[-7.17,-6.52]$ \\
Qwen3 4B & .01 & -- & -- & $+1.17$ & $-9.19$ & -- & $-0.05$ & $-1.34$ & -- & -- \\
Qwen3 8B & .00 & $+9.00$ & $+1.07$ & $+0.87$ & $-10.51$ & $-2.94$ & $+0.05$ & $-1.27$ & $-1.06$ & $-9.44$ $[-9.81,-9.08]$ \\
\bottomrule
\end{tabular}
\caption{Blocking selected prompt spans for the \heldout{} form (item sets 0 and 1, 320 items, bfloat16; $^{\dagger}$one set, 160 items). The body and Figure~\ref{fig:knockout} quote float32 values on item sets 2 and 3 (supplement table). $m$ (unblocked) is the full-prompt correct-minus-copied log-probability in nats; every other entry is the change in $m$ after blocking the named span. \emph{Closest example}: only its ending tokens (\emph{ending}) or its two whole lines (\emph{two lines}). \emph{Question words}: the word naming the requested value (\emph{asked}) or the other value (\emph{other}). \emph{Controls}: two unrelated single-feature lines (\emph{unrelated}), the two single-feature lines (\emph{single-feature lines}) and only their ending tokens (\emph{their endings}). $D$ is the magnitude for \emph{ending} minus that for \emph{asked}, with a 95\% item-bootstrap interval. -- marks a span not run (Qwen3 4B's closest-ending block) or not tabulated here. $^{\ddagger}$Run added after the model set was fixed; float32 and bfloat16 agree to .02 nats for this model. Four of the 17 models of Table~\ref{tab:fingerprint} (Pythia 410M and 6.9B, Qwen2.5 1.5B and 3B) were not run.}
\label{tab:knockout}
\end{table*}

%% file: tables/deletion_final_app.tex
\begin{table}[htbp]
\centering\footnotesize
\setlength{\tabcolsep}{1.5pt}
\begin{tabular}{@{}lrrrr@{}}
\toprule
Model & full $m$ & \shortstack{closest\\example} & \shortstack{correct-ending\\lines} & \shortstack{unrelated\\lines} \\
\midrule
Pythia 1.4B & $-2.08$ & $+3.20$ & $-0.53$ & $-0.13$ \\
Pythia 2.8B & $-1.92$ & $+3.30$ & $-0.92$ & $+0.05$ \\
Qwen3 0.6B & $-0.19$ & $+3.52$ & $-1.28$ & $-0.40$ \\
OLMo-2 7B & $+2.31$ & $+1.66$ & $-1.18$ & $-0.49$ \\
OLMo-2 13B & $+3.53$ & $+1.59$ & $-0.95$ & $-0.77$ \\
Qwen3 1.7B & $+3.87$ & $+1.51$ & $-1.36$ & $-0.44$ \\
Gemma-3 4B & $+3.72$ & $+1.20$ & $-1.33$ & $-0.89$ \\
Qwen3 8B & $+8.98$ & $-0.19$ & $-1.52$ & $-0.50$ \\
\bottomrule
\end{tabular}
\caption{Deletion effects on item sets 0 and 1, 160 items each. Full $m$ is the unblocked correct-minus-copied log-probability in nats. Columns 3--5 give the change in $m$ after deleting the closest example's two lines, the two single-feature lines defining the correct second ending, or two unrelated lines of matched length.}
\label{tab:deletion_final}
\end{table}

%% file: tables/p7a_stagea_app.tex
\begin{table*}[htbp]
\centering\setlength{\tabcolsep}{3pt}\renewcommand{\arraystretch}{1.0}
\setlength{\tabcolsep}{2.5pt}\scriptsize
\begin{tabular}{@{}lrrrrrl@{}}
\toprule
Released model & \shortstack{correct ranks\\first (of 320)} & \shortstack{writes\\correct} & $P(\mathrm{copy})$ & $\Delta$ closest ending & $D$ & source \\
\midrule
Marin-8B-Base$^{\dagger}$ & 283 (0.88) & .62 & .02 & $+1.52$ $[1.31, 1.73]$ & $-6.19$ & \citet{marincommunity2025marin8b} \\
Comma-v0.1-1T$^{\dagger}$ & 277 (0.87) & .50 & .06 & $+1.46$ $[1.23, 1.69]$ & $-6.40$ & \citet{kandpal2025commonpile} \\
SmolLM3-3B$^{\dagger}$ & 271 (0.85) & .54 & .10 & $+1.74$ $[1.51, 1.96]$ & $-5.10$ & \citet{huggingface2025smollm3} \\
OLMo-3-7B$^{\dagger}$ & 265 (0.83) & .43 & .05 & $+1.63$ $[1.40, 1.85]$ & $-6.24$ & \citet{olmo3} \\
\midrule
Apertus-8B-2509 & 136 (0.42) & .15 & .23 & $+2.13$ $[1.91, 2.36]$ & $-3.56$ & \citet{swissai2025apertus8b} \\
\midrule
OpenLLaMA-7B-v2$^{\dagger}$ & 55 (0.17) & .06 & .55 & $+2.98$ $[2.78, 3.18]$ & $-0.20$ & \citet{openlm2023openllama7bv2} \\
SmolLM2-1.7B$^{\dagger}$ & 38 (0.12) & .05 & .64 & $+3.92$ $[3.71, 4.12]$ & $+1.13$ & \citet{allal2025smollm2} \\
Lucie-7B & 30 (0.09) & .03 & .72 & $+4.51$ $[4.31, 4.71]$ & $+2.22$ & \citet{openllmfrance2025lucie7b} \\
YuLan-Mini (2.4B) & 27 (0.08) & .06 & .68 & $+4.24$ $[4.02, 4.45]$ & $+0.92$ & \citet{hu2024yulanmini} \\
RedPajama-INCITE-7B-Base & 26 (0.08) & .03 & .64 & $+3.84$ $[3.65, 4.04]$ & $+1.14$ & \citet{together2023redpajama7b} \\
Viking-7B & 18 (0.06) & .02 & .61 & $+3.55$ $[3.29, 3.78]$ & $+0.53$ & \citet{lumiopen2024viking7b} \\
BLOOM-7b1 & 14 (0.04) & .02 & .76 & $+2.96$ $[2.75, 3.17]$ & $+0.95$ & \citet{bigscience2024bloom} \\
INTELLECT-1 (10B) & 11 (0.03) & .01 & .74 & $+2.83$ $[2.62, 3.03]$ & $+0.77$ & \citet{jaghouar2024intellect} \\
FineWeb-Edu ablation model (1.82B) & 3 (0.01) & .01 & .94 & $+7.35$ $[7.03, 7.64]$ & $+6.48$ & \citet{huggingfacefw2024finewebeduablation} \\
SmolLM2-360M & 3 (0.01) & .00 & .90 & $+4.52$ $[4.30, 4.76]$ & $+3.80$ & \citet{allal2025smollm2} \\
TinyLlama-1.1B & 2 (0.01) & .01 & .90 & $+4.90$ $[4.69, 5.11]$ & $+4.11$ & \citet{zhang2024tinyllama} \\
\bottomrule
\end{tabular}
\caption{Screen of 16 released models (320 items), sorted by the number of items on which the correct form ranks first. Rules separate high (at least 160 correct), partial (Wilson lower bound above $.25$ but fewer than 160 correct) and low (neither). ``$\Delta$ closest ending'' is the change in correct-minus-copied log-probability after blocking that ending, with its 95\% interval; $D$ subtracts the magnitude after blocking the requested value. $^{\dagger}$The six runs followed across their saved intermediate models: the four high models and two contrast series.}
\label{tab:p7a_stagea}
\end{table*}

%% file: tables/crossings_app.tex
\begin{table*}[htbp]
\centering\scriptsize
\setlength{\tabcolsep}{4.0pt}\renewcommand{\arraystretch}{1.0}
\begin{tabular}{@{}lrrrl@{}}
\toprule
Trajectory & $P(\mathrm{copy})<.5$ & $D<0$ & \shortstack{correct ranks\\first $>.25$} & \shortstack[l]{fitted accuracy\\threshold crossing} \\
\midrule
OLMo-2-7B & 630B & 630B & 1.7T & 1.79T \\
OLMo-2-13B & 2.5T & 336B & 2.5T & 2.2T \\
OLMo-2-1B & never & never & never & -- \\
OLMo-7B (0424) & 838B & 838B & 2.0T & 1.6T \\
Pythia-1.4B & never & never & never & -- \\
Pythia-2.8B & never & never & never & -- \\
Pythia-6.9B & never & never & never & -- \\
Pythia-12B & never & never & never & not reached (last ckpt 300B) \\
OLMoE-1B-7B & never & never & never & not reached (last ckpt 5T) \\
Amber-7B (LLM360) & never & never & never & not reached (last ckpt 1.3T) \\
Pythia-1.4B-deduped & never & never & never & -- \\
OLMo-3-7B & 1.3T & 600B & 1.3T & 1.0T \\
Marin-8B-Base & $\leq$2.7T & $\leq$2.7T & $\leq$2.7T & -- \\
SmolLM3-3B & 1.9T & 1.9T & 4.1T & 7.6T \\
SmolLM2-1.7B & never & never & never & -- \\
Comma-v0.1-1T & 818B & 545B & 818B & 660B \\
OpenLLaMA-7B-v2$^{\dagger}$ & never & 460k & never & -- \\
\bottomrule
\end{tabular}
\caption{First sampled checkpoint crossing each threshold on 320 independently generated items. The fitted column is the logistic crossing for the fraction of prompts on which the correct form ranks first (Appendix~\ref{app:onsets}); ``not reached'' marks a fitted crossing beyond the last observed checkpoint. $\leq$: Marin-8B-Base already exceeds the threshold at its first public checkpoint. OLMo-2-7B's fitted crossing is the re-test value, 1.79T (Appendix~\ref{app:confirm}). A series may return after its first crossing. $^{\dagger}$OpenLLaMA positions are optimizer steps rather than tokens. The six ordering criteria per run are visible in Figure~\ref{fig:crossings}.}
\label{tab:crossings}\label{tab:trajdesc}
\end{table*}

%% file: tables/stage2_app.tex
\begin{center}
\setlength{\tabcolsep}{6pt}
\setlength{\tabcolsep}{2.5pt}\scriptsize
\begin{tabular}{@{}lcc@{}}
\toprule
OLMo-2-7B checkpoint & \shortstack{correct form\\ranks first} & \shortstack{writes\\correct} \\
\midrule
end of stage 1 (3.9T) & .53 & .17 \\
first stage-2 run, 5B more tokens & .64 & .22 \\
released model (parameter average) & .69 & .30 \\
\bottomrule
\end{tabular}
\end{center}

%% file: tables/confirm_finals_app.tex
\begin{table*}[htbp]
\centering\scriptsize
\setlength{\tabcolsep}{1.5pt}\renewcommand{\arraystretch}{1.0}
\begin{tabular}{@{}lccrrrrlccc@{}}
\toprule
Model & $P(\mathrm{copy})$ [95\% interval] & \shortstack{full\\prompt} & \shortstack{closest\\ending} & \shortstack{requested\\value} & \shortstack{other\\value} & \shortstack{unrelated\\lines} & \multicolumn{1}{c}{$D$ [95\% interval]} & \shortstack{$P(\mathrm{copy})$\\predicted side\\$\geq .70 / \leq .30$} & \shortstack{$D$ sign\\as predicted} & \shortstack{unrelated 95\%\\interval within\\$\pm .3$ nats} \\
\midrule
Pythia 1.4B & .81 [.77, .83] & $-2.73$ & $+5.33$ & $-2.06$ & $+1.47$ & $+0.24$ & $+3.27$ $[+2.86,+3.68]$ & \checkmark & \checkmark & $\times$ \\
Pythia 2.8B & .77 [.74, .79] & $-2.08$ & $+4.63$ & $-2.26$ & $+1.05$ & $+0.16$ & $+2.38$ $[+2.10,+2.64]$ & \checkmark & \checkmark & \checkmark \\
Pythia 12B & .85 [.82, .87] & $-3.20$ & $+5.35$ & $-1.63$ & $+1.83$ & $+0.17$ & $+3.72$ $[+3.32,+4.08]$ & \checkmark & \checkmark & $\times$ \\
OLMo-2 1B & .57 [.53, .60] & $-0.43$ & $+2.80$ & $-3.40$ & $+1.00$ & $+0.11$ & $-0.60$ $[-0.98,-0.23]$ & -- & -- & -- \\
Qwen3 0.6B & .53 [.50, .56] & $-0.15$ & $+2.88$ & $-4.39$ & $+0.78$ & $+0.04$ & $-1.52$ $[-1.85,-1.17]$ & -- & -- & -- \\
\midrule
OLMo-2 7B & .24 [.21, .27] & $+2.54$ & $+2.08$ & $-5.54$ & $-0.77$ & $+0.10$ & $-3.46$ $[-3.93,-3.02]$ & \checkmark & \checkmark & \checkmark \\
OLMo-2 13B & .16 [.13, .19] & $+3.74$ & $+1.55$ & $-7.07$ & $-1.65$ & $-0.13$ & $-5.52$ $[-6.09,-5.01]$ & \checkmark & \checkmark & \checkmark \\
Qwen3 1.7B & .10 [.08, .12] & $+3.97$ & $+1.83$ & $-7.07$ & $-0.15$ & $-0.06$ & $-5.25$ $[-5.65,-4.81]$ & \checkmark & \checkmark & \checkmark \\
Gemma-3 4B & .11 [.09, .13] & $+4.17$ & $+2.64$ & $-7.10$ & $-1.54$ & $-0.08$ & $-4.46$ $[-4.90,-4.04]$ & \checkmark & \checkmark & \checkmark \\
Qwen2.5 7B & .02 [.01, .02] & $+5.75$ & $+0.98$ & $-7.99$ & $-1.74$ & $-0.04$ & $-7.02$ $[-7.36,-6.64]$ & \checkmark & \checkmark & \checkmark \\
Qwen3 8B & .00 [.00, .00] & $+8.89$ & $+0.87$ & $-10.41$ & $-2.96$ & $+0.05$ & $-9.54$ $[-9.92,-9.16]$ & \checkmark & \checkmark & \checkmark \\
\bottomrule
\end{tabular}
\caption{Re-test on item sets 2 and 3 (320 items per model, bfloat16). $P(\mathrm{copy})$ and $D$ have 95\% item-bootstrap intervals, resampling within each set. Blocking columns give changes in $m$ (nats); $D$ is the closest-ending change minus the magnitude of the requested-value change. The last columns test the predicted probability side, sign of $D$, and unrelated-line interval. The two intermediate models have no directional prediction. The body quotes float32 re-scores of these same items: Pythia $P(\mathrm{copy})$ is lower by up to $.06$ and $D$ by up to $1.8$ nats.}
\label{tab:confirm_finals}
\end{table*}

%% file: sections/highlight/09b_part7b_h.tex
\section{Continued pretraining with synthetic documents and attention penalties (360M)}
\label{app:part7b}
\begin{figure*}[t]
\centering
\includegraphics[width=\textwidth]{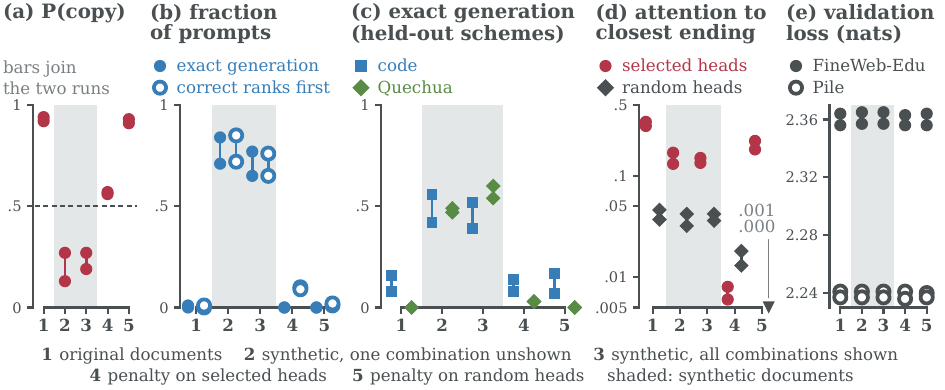}
\caption{SmolLM2-360M after about 1B additional tokens. Conditions: original documents, one-unshown and all-shown synthetic documents, and penalties on selected or random heads, two runs per condition as joined markers. (a) $P(\mathrm{copy})$, (b) exact generation and correct-ranks-first, (c) code and Quechua exact generation, (d) selected-head and random-head attention to the closest ending, and (e) FineWeb-Edu and Pile validation loss. Both synthetic conditions cross $D<0$ and the accuracy criterion at step 32 (about 50M tokens). The other conditions never cross within 1B.}
\label{fig:p7b_final}
\end{figure*}

Five conditions continue pretraining SmolLM2-360M for about 1B tokens, two runs each: original documents, two synthetic-document variants and two attention penalties (Section 6.1; Figure~\ref{fig:p7b_final}).

\paragraph{Design.} All conditions share the starting checkpoint, compute budget and data order.
\begin{itemize}\setlength{\itemsep}{0pt}
\item Base: \texttt{SmolLM2-360M} at \texttt{step-1600000} (2.52 trillion tokens), continued on the public proxy corpus \texttt{smollm-corpus} \citep{allal2025smollm2} for 636 steps, about 1 billion tokens.
\item Synthetic documents: every 33rd packed sequence (3\,\%, about 30 million tokens) is replaced by documents that define endings. One-unshown documents leave one combination unshown and give its answer. All-shown documents show every requested combination on another stem. Training stems, endings and templates are disjoint from the evaluation inventories, and neither condition contains code systems.
\item Attention penalties: original documents plus a penalty on attention from the answer position to the closest ending, on either the ten heads with the largest direct contribution to the copied answer in the base model or ten random heads in the same layers (Appendix~\ref{app:part6}).
\item Measures: $P(\mathrm{copy})$, correct-ranks-first, exact generation and $D$, code and Quechua scores, three external evaluations defined in the supplement, FineWeb-Edu and Pile validation losses, and the penalized attention.
\item Crossings: a condition's $D$ turns negative at the first checkpoint with mean $D<0$ and $P(\mathrm{copy})\leq .55$, and it reaches the accuracy criterion at the first with correct-ranks-first at least $.50$ and exact generation at least $.15$.
\end{itemize}

\paragraph{Comparisons.} An intervention counts as earlier only if it crosses before its control, or crosses when the control never does. Transfer to code and Quechua requires a gain of at least $.05$ on both measures of both tasks in both runs, each with a positive 95\% paired item-bootstrap interval, and both validation losses within $+.02$ nats of original documents.

\paragraph{Outcome.} Both synthetic conditions cross both thresholds at step 32, about 50M tokens, so they tie and neither counts as earlier (Figure~\ref{fig:p7b_final}). At the end the all-shown runs have $P(\mathrm{copy})$ of $.27/.19$ and exact generation of $.65/.77$. The one-unshown runs have $.27/.13$ and $.71/.84$. The one-unshown condition meets the transfer and loss criteria against original documents but not against all-shown documents. The selected-head penalty drives the penalized attention from $.310/.341$ to $.008/.006$, yet $P(\mathrm{copy})$ stays at $.56/.57$, $D$ at $+1.01/+1.60$, and exact generation at zero. Original documents and both penalty conditions reach neither threshold within 1B tokens, and the selected-head penalty does not reach the transfer threshold. Practicing an unshown combination is therefore not isolated as the cause, and reducing the selected heads' attention to the closest ending does not create composing.

%% file: sections/highlight/09e_part10_h.tex
\section{Continued pretraining with synthetic documents and mined passages (1.7B)}
\label{app:part10}
\begin{figure*}[t]
\centering
\includegraphics[width=\textwidth]{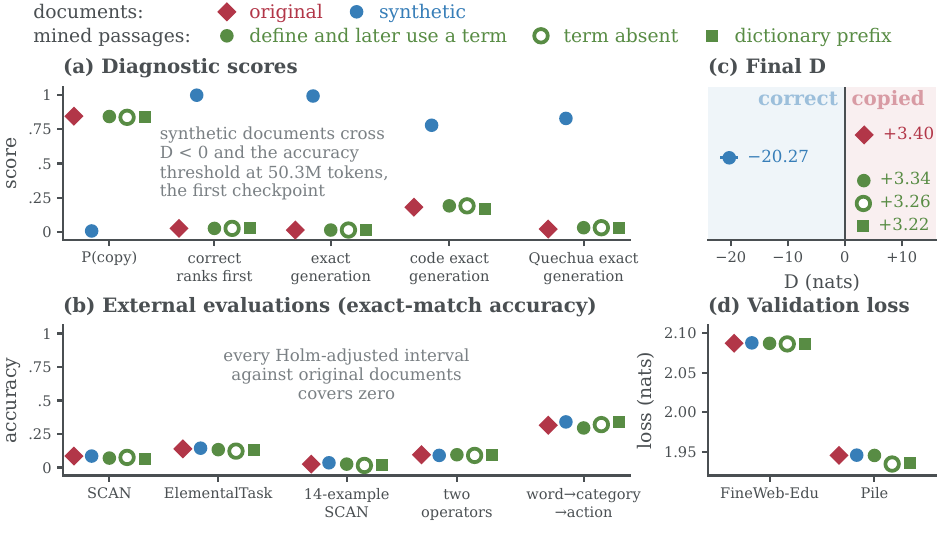}
\caption{SmolLM2-1.7B after 1B additional tokens. Series are original documents, 3\% synthetic documents, mined define-and-use passages, mined passages with the term absent from the reference vocabulary, and mined passages with a dictionary prefix. (a) $P(\mathrm{copy})$, correct-ranks-first, exact generation, and code and Quechua exact generation. (b) Exact-match accuracy on SCAN, ElementalTask, 14-example SCAN, two operators, and word-to-category-to-action. No external-task change is significant. (c) Final $D$. (d) Validation losses.}
\label{fig:p10all_final}
\end{figure*}

Five conditions continue pretraining SmolLM2-1.7B for 1B tokens: plain continuation, 3\% synthetic documents, and three mined-passage variants (Sections 6.1--6.2; Figures~\ref{fig:p10all_final} and~\ref{fig:lever}).

\paragraph{Design.} All conditions share the base checkpoint and training schedule.
\begin{itemize}\setlength{\itemsep}{0pt}
\item Base: \texttt{SmolLM2-1.7B-intermediate-\allowbreak checkpoints} at 2.097 trillion tokens \citep{allal2025smollm2}.
\item Schedule: 477 steps of 2.097 million tokens (1B total) at learning rate $5\times10^{-4}$, identical rows across conditions, every 33rd row ($3.03\,\%$) reserved for replacement, and 21 checkpoints.
\item Conditions: original documents, the synthetic documents of Appendix~\ref{app:part7b}, mined passages that define and later use a term, mined passages whose term is absent from a fixed reference vocabulary, and the same passages with a dictionary-format prefix. The all-shown control was not repeated at this scale for lack of compute (Appendix~\ref{app:disclosures}).
\item Measures: the diagnostic ($n=160$) with its code and Quechua versions, five external tasks ($n=300$ for the first, $n=200$ for the others), and FineWeb-Edu and Pile loss \citep{penedo2024fineweb,gao2020pile}. The external tasks are in-context SCAN (add-primitive-jump split) and a 14-example SCAN-style task \citep{lake2018generalization}, in-context ElementalTask \citep{liu2026curriculum}, nested use of two defined operators, and an unshown word-to-action mapping through word-to-category and category-to-action definitions. None passes the pre-set qualification threshold.
\item Crossings: a condition's $D$ turns negative when $D<0$ with $P(\mathrm{copy})\leq .55$, and it reaches the accuracy criterion when correct-ranks-first is at least $.50$ and exact generation at least $.15$.
\end{itemize}

The fixed comparisons require each intervention to cross while original documents do not and to keep both validation losses within $+.02$ nats. Mined passages must cross no later than synthetic documents. External-task changes are evaluated with Holm-adjusted item-bootstrap intervals but cannot establish qualified transfer.

\paragraph{Outcome.} After 1B tokens, synthetic-document training has $P(\mathrm{copy})=.006$ and exact generation $.994$, while plain and mined-passage runs keep $P(\mathrm{copy})$ near $.84$ and exact generation at $.013$ (Figure~\ref{fig:p10all_final}). No external task changes by more than $.03$ under any condition. Synthetic documents cross both thresholds at the first checkpoint, 50.3 million tokens. The other four conditions never cross within 1B.\footnote{The synthetic-document comparison meets its criterion only after a Quechua scoring fix. The first analysis read a missing correctness field and scored every Quechua item 0 (Appendix~\ref{app:disclosures}).} FineWeb-Edu loss changes by at most $+.0005$ nats, and Pile changes range from $-.0110$ to $+.0003$. The comparisons are between whole training conditions and do not isolate which property of the documents matters.

%% file: tables/doc_examples.tex
\begin{table*}[t]\centering\footnotesize
\begin{tabular}{@{}p{0.23\textwidth}p{0.22\textwidth}p{0.49\textwidth}@{}}
\toprule
Modification type & Original document (excerpt) & Exercise built from it: definitions $\rightarrow$ query $\rightarrow$ answer \\
\midrule
Symbols as labels for passages, the rule shown only by demonstrations (type b) & ``\ldots{} Import databases into GIS software. \ldots'' (document dd98ff1b3ceb) & ``Passage: Import databases into GIS software. Label: VENTRIX'' ``Passage: Identify locations and spatial extent of major climate hazards. Label: HOLPREN'' $\rightarrow$ ``What is the label for: 'Data sources for assessing conditions and trends.'?'' $\rightarrow$ \textbf{HOLPREN} \\
\addlinespace
A stated mapping that contradicts the document's own names (type c) & ``\ldots{} Clemens's research is focused on social movements, organizations, and American pol \ldots'' (document 72d352b5830b) & ``California is called Wisconsin, Wisconsin is called Washington, and Washington is called California.'' $\rightarrow$ ``what is the renamed version of the state that the original document calls 'California'?'' $\rightarrow$ \textbf{Wisconsin} \\
\addlinespace
A small invented vocabulary with grammar rules (type e) & ``\ldots{} The Brewarrina Aboriginal Cultural Museum stands at a site which has been of great importance to Aboriginal people \ldots'' (document a7c757380862) & ``mara = museum'' ``kelu = story'' ``silka = respect'' ``adding the suffix -oz to a noun marks more than one of that thing (plural)'' $\rightarrow$ ``Translate into Bralin: 'respect''' $\rightarrow$ \textbf{silka} \\
\addlinespace
Private codes bound to the document's entities (type g) & ``\ldots{} Printable PDF Worksheets \ldots'' (document 5875e15ba879) & ``The engineering team referred to the printable worksheet generator internally as Rontavex.'' $\rightarrow$ ``Which nickname is bound to the attribute describing syncing marked statuses back after a device switch?'' $\rightarrow$ \textbf{Dufaline} \\
\addlinespace
A mapping where the shown answer is wrong and only the given rule is right (type h) & ``\ldots{} Ensuring your safety while camping should be a top priority. \ldots'' (document f99176a67baa) & ``Input: campsite selection distance from water $\rightarrow$ Output: RV3'' ``Input: wildlife encounter bear-proof containers $\rightarrow$ Output: RV2'' $\rightarrow$ ``Input: campsite selection near water signal $\rightarrow$ Output:'' $\rightarrow$ \textbf{RV3} \\
\addlinespace
\bottomrule
\end{tabular}
\caption{One real example per modification type: a passage from a real document (left) and the definition-use exercise a language model built from it (right; definitions, then the query and its answer), each verified by a second model before training.}
\label{tab:doc_examples}
\end{table*}